\documentclass[letterpaper, paper,11pt]{AAS}		

\usepackage{graphicx}
\usepackage{subfig}
\usepackage{bm}
\usepackage{amsmath}
\usepackage{subfig}
\usepackage{float}
\usepackage[colorlinks=true, pdfstartview=FitV, linkcolor=black, citecolor= black, urlcolor= black]{hyperref}
\usepackage{overcite}
\usepackage{footnpag}			      	
\usepackage{xcolor}
\usepackage{booktabs}
\usepackage{multirow}
\usepackage{bbm}
\usepackage{hyperref}

\PaperNumber{22-638}

\begin{document}

\title{You Only Crash Once: Improved object detection for real-time, sim-to-real hazardous terrain detection and classification for autonomous planetary landings}

\author{Timothy~Chase~Jr\thanks{Graduate Student, Department of Computer Science \& Engineering, University at Buffalo, The State University of New York, Buffalo NY 14260, \href{mailto:tbchase@buffalo.edu}{\color{blue}tbchase@buffalo.edu}},
Chris~Gnam\thanks{Graduate Student, Department of Mechanical \& Aerospace Engineering, University at Buffalo, The State University of New York, Buffalo NY 14260, \href{mailto:crgnam@buffalo.edu}{\color{blue}crgnam@buffalo.edu}},
John~Crassidis\thanks{SUNY Distinguished Professor and Samuel P.~Capen Chair Professor, Department of Mechanical \& Aerospace Engineering, University at Buffalo, The State University of New York, Buffalo NY 14260, Fellow AIAA, \href{mailto:johnc@buffalo.edu}{\color{blue}johnc@buffalo.edu}}, 
Karthik~Dantu\thanks{Associate Professor, Department of Computer Science \& Engineering, University at Buffalo, The State University of New York, Buffalo NY 14260, \href{mailto:kdantu@buffalo.edu}{\color{blue}kdantu@buffalo.edu}}}

\maketitle{}


\begin{abstract}
The detection of hazardous terrain during the planetary landing of spacecraft plays a critical role in assuring vehicle safety and mission success. A cheap and effective way of detecting hazardous terrain is through the use of visual cameras, which ensure operational ability from atmospheric entry through touchdown. Plagued by resource constraints and limited computational power, traditional techniques for visual hazardous terrain detection focus on template matching and registration to pre-built hazard maps. Although successful on previous missions such as the landing of the Mars Perseverance Rover, this approach is restricted to the specificity of the templates and limited by the fidelity of the underlying hazard map, which both require extensive pre-flight cost and effort to obtain and develop. It would thus be more beneficial to have a system capable of a general perception towards a wide range of hazardous terrain. Terrestrial systems that perform a similar task in applications such as autonomous driving utilize state-of-the-art deep learning techniques to successfully localize and classify navigation hazards. Advancements in spacecraft co-processors aimed at accelerating deep learning inference enables the application of these methods in space for the first time. In this work, we introduce \textit{You Only Crash Once} (YOCO), a deep-learning based visual hazardous terrain detection and classification technique for autonomous spacecraft planetary landings. Through the use of unsupervised domain adaptation we tailor YOCO for training by simulation, removing the need for real-world annotated data and expensive mission surveying phases. We further improve the transfer of representative terrain knowledge between simulation and the real-world through visual similarity clustering. We demonstrate the utility of YOCO through a series of terrestrial and extraterrestrial simulation-to-real experiments, and show substantial improvements towards the ability to both detect and accurately classify instances of planetary terrain.

\end{abstract}

\section{Introduction}
When spacecraft are tasked with landing on the surface of other planets such as Mars, scientific objectives often guide the spacecraft to a landing site within close proximity of terrain that is hazardous to the spacecraft. Terrain such as canyons, cliffs, dunes, rock fields, and craters must be identified as quickly as possible during the entry, descent, and landing (EDL) maneuver in order to ensure spacecraft safety to the highest degree of precision possible. Terrain Relative Navigation (TRN) plays an important role in the EDL process by detecting terrain landmarks during descent, and using these detections to estimate a vehicle position fix relative to a pre-determined map of the landing site. Traditionally, vision-based systems have been used for detecting these landmarks from real-time image frames captured from a downward facing camera on the landing spacecraft, which are then matched to the underlying map through template matching approaches. By identifying hazardous terrain \textit{a priori} in the map, TRN can be leveraged to detect and maneuver the spacecraft away from such regions. This approach has found success on previous missions such as the landing of the Mars Perseverance Rover~\cite{m2020lvs}. However, through the template matching approach, substantial pre-flight mission cost and effort was devoted towards analyzing the landing site for hazardous terrain, and integrating these hazards into the relative navigation map. It would be more beneficial to have a vision system capable of a general perception towards hazardous terrain, not limited by knowledge only gained through pre-flight reconnaissance and mission surveying phases and without the reliance on a pre-determined navigation map. General terrain perception further leads to enhanced semantic understanding of the environment, as well as more efficient and accurate mapping capability~\cite{zaki_semantic, zaki_semantic2}.

Informed by the growing use of computer vision within real-time safety-critical scenarios such as autonomous driving, this paper explores the application of terrestrial state-of-the-art object detection methods to the in-situ hazardous terrain identification problem for a spacecraft landing scenario within a planetary environment. We make the following contributions in this work: (i) We adapt \textit{You Only Look Once} (YOLO)~\cite{yolo}, a widely popular, real-time object detection algorithm towards the detection and classification of planetary terrain.  (ii) With no labelled training data readily available, we augment the YOLO architecture with domain adaptation methods that enable training through simulation, effectively removing the dependency on \textit{a priori} hazardous terrain knowledge needed during the landing operation. (iii) As classes of terrain are relatively diverse, we integrate visual similarity-based clustering into the domain adaptation framework, simplifying the problem of object alignment and enabling sufficient robustness to intra-class appearance differences in the terrain such as size, shape, and texture.



\section{Terrestrial Object Detection Techniques}
Object detection has been a revolutionary application of computer vision in recent years. The powerful ability to classify and localize objects of interest from a single camera has many applications, especially in safety-critical systems such as autonomous navigation. Recent state-of-the-art object detection systems have been dominated by the field of deep learning, where architectures such as Convolutional Neural Networks (CNNs) have been applied with great success. More generally, systems that utilize deep learning for object detection are divided into two categories: two-stage and one-stage detectors. Two-stage detectors tackle the object detection problem through two sequential steps, which include the generation of proposed object regions on the image and a prediction of object classification for each proposal. In contrast, one-stage detectors instead focus on the regression of bounding box coordinates from an input image which greatly simplifies the model architecture. A single network is used to extract image features, produce probabilities of object existence, and regress bounding box coordinates for each object in one forward pass. This process allows one-stage detectors to achieve much higher running times than two-stage detectors at the cost of slightly lower localization accuracy.




\subsection{Faster R-CNN}
The notion of two-stage detectors was first introduced by the Faster R-CNN~\cite{fasterrcnn}  architecture. The pipeline included two modules, the first being a region proposal network (RPN) and the second being the Fast R-CNN~\cite{fastrcnn} network. 
As part of the training process regions of interest are both proposed and refined through the RPN which acts as an attention mechanism to highlight ``where to look'' in the image. A pre-trained network is used as an initial feature extractor to seed the RPN, which uses pre-defined shapes and sizes (called anchor boxes) as a starting point to aid the process of determining if the input image contains any objects. 
For each region proposal the RPN also outputs a binary class prediction for each, indicating the presence of an object or not. This ``\textit{objectness}'' score helps eliminate regions before supplying them to the further computationally intensive layers of the Fast R-CNN model. The surviving regions output by the RPN are fed to the Fast R-CNN module, which acts as the actual detection component. With a set of region proposals output by the RPN, the Fast R-CNN detector learns to fine-tune region coordinates and predict class labels. The RPN and Fast R-CNN components of the unified Faster R-CNN model are trained alternatively, in which Fast R-CNN is used to first initialize the RPN, and the subsequent proposals are used to train Fast R-CNN. After which the process is iterated. Despite state-of-the-art performance in object detection and localization accuracy benchmarks, the runtime of Faster R-CNN only reaches a frame rate of roughly 5fps on modern GPUs, which has been a large contributing factor towards it's exclusion in safety critical embedded systems.



\subsection{You Only Look Once (YOLO)}
The YOLO family of deep-learning models has ushered in the era of real-time object detection with excellent performance. As a one-shot detector, YOLO uses a single neural network that's trained end to end to jointly predict bounding boxes and class labels for each box in a single forward pass. Without the use of a separate region proposal network, YOLO models object detection as a regression problem. This is achieved by first splitting the input image into a grid of cells, where each cell is responsible for predicting a set number of bounding boxes and a confidence score for each. 
For boxes that believe no object exists, their confidence scores are driven close to zero. Conversely, a bounding box with a confidence score of one points to a high certainty of object existence within that cell. Each cell is also responsible for predicting the class of the object it seeks to localize. Despite producing multiple bounding box and respective confidence score predictions, the cells produce only \textbf{one} single object classification. This is a limitation of the model which degrades localization accuracy for objects that are close together (or overlapping) in the image. 
Further, YOLO produces many duplicate object predictions as most nearby cells to an object predict and attempt to localize that object. Non-max suppression is used to reduce and/or eliminate these duplicate predictions, resulting in a final best-fitting and highest confidence bounding box around a given object. 
Attributed to the single network architecture, standard YOLO models operate at 45fps while more speed-optimized versions operate in excess of 145fps. This makes YOLO very attractive for use in safety critical embedded systems, particularly in navigation where vehicles need to make corrective maneuvers on the fly in real-time at high speeds.

\begin{figure*}
	\centering
	\includegraphics[width=.8\linewidth]{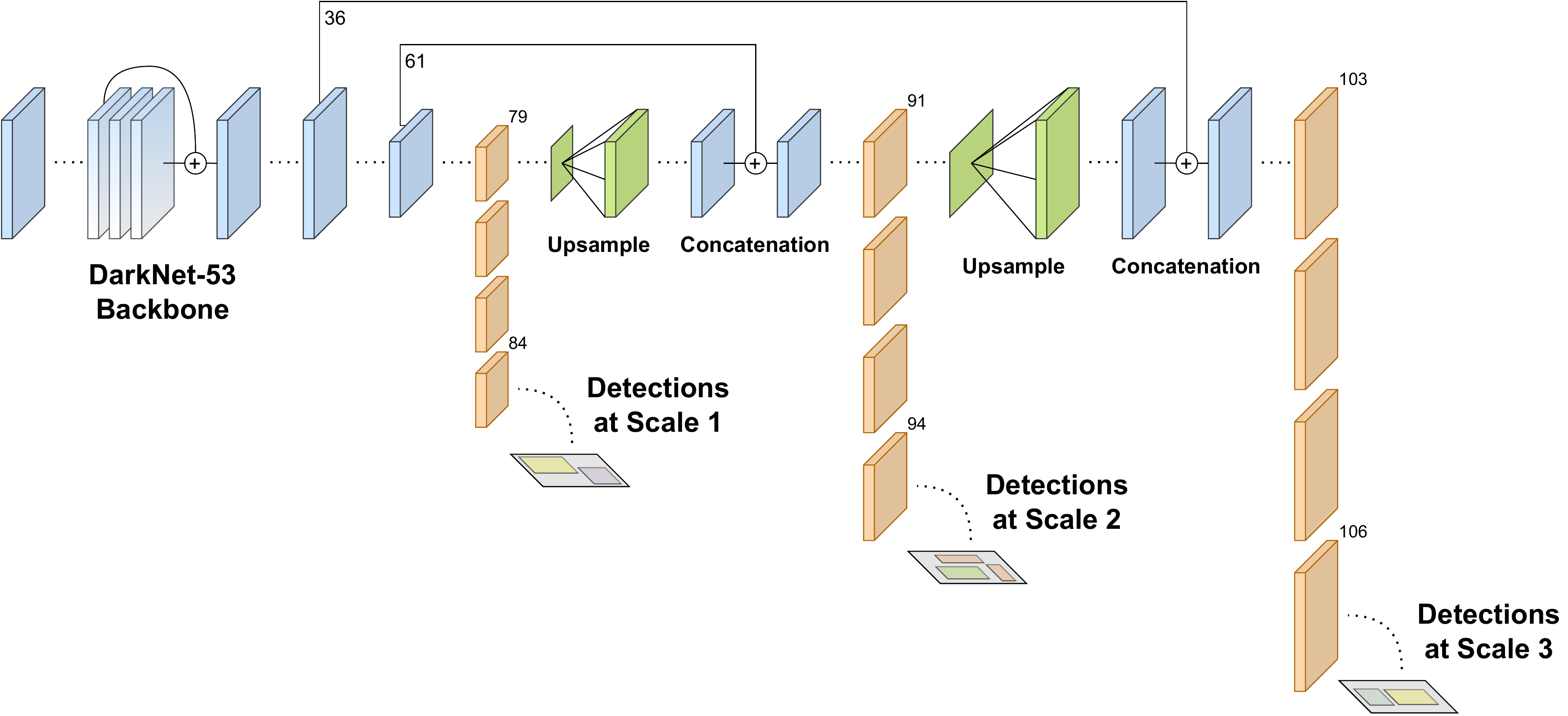}
	\caption{Network architecture of YOLOv3~\cite{yolov3}.}
	\label{fig:yolo_arch}
	\vspace{-20pt}
\end{figure*}

The YOLO architecture consists of two key components, a backbone and detection head. The backbone contains a set of pre-trained CNN layers meant to serve as the initial feature extraction component of the model. Features extracted from the backbone are fed to the detection head of the network, which is responsible for regressing the outputs of bounding box coordinates and respective confidence score and class label. YOLOv3~\cite{yolov3} is a recent version of YOLO that improves upon it's initial design. It leverages DarkNet-53 as a backbone feature extractor which contains 53 convolutional layers (opposed to the original 24) along with residual and skip connections to improve performance. YOLOv3 also improves the scale at which detections can be made by incorporating three separate detection heads. The model architecture of YOLOv3 is shown in~\autoref{fig:yolo_arch}.

\section{Object Detection Challenges in Space Environments}
\subsection{Processing Limitations}
Although YOLO has found extraordinary success in a wide-range of terrestrial scenarios, there are several challenges when applying it towards spacecraft applications.
Perhaps the biggest challenge is the amount of processing power the algorithm requires. Spacecraft flight computers and information processing systems notoriously lag behind the latest terrestrial state-of-the-art, mostly due to size, radiation tolerance, and power footprint requirements. Due to this, spacecraft performing safety-critical detections at high rate during a landing maneuver have been unable to reap the benefits that the latest in computer vision research brings. For similarly resource constrained applications on Earth, Google has developed the Coral Edge Tensor Processing Unit (TPU) aimed to accelerate deep learning model inference on embedded devices which can bring model inference times down to the millisecond range. 
NASA’s Goddard Space Flight Center (GSFC) has recently designed an embedded spacecraft co-processor, the SpaceCube LEARN (SC-LEARN)~\cite{sclearn}, adapting these TPUs. SC-LEARN has completed extensive environmental and radiation testing (proton, heavy-ion, total ionizing dose) and is at TRL-6 at the time of writing. With the addition of TPUs now available as an in-flight hardware accelerator, newer generations of deep learning models can be deployed in space for real-time safety-critical operations.


\subsection{Training Data Availability and Model Performance}
Another obstacle preventing the realization of using learning-based vision for space applications is the lack of labelled training data. YOLO being a supervised learning system requires training images (in thousands) annotated with bounding box locations for all objects along with respective class labels. For many space related applications, such labeled data is extremely hard to come by or completely non-existent. Limited data availability is a problem shared with applications in autonomous robotics where labelled data is similarly sparse. A traditional workaround in these terrestrial robotics applications is to leverage photo-realistic simulations to automatically generate and annotate the needed training data.

Models trained on simulated data when used for operation in the real-world suffer from underlying data distribution differences with varying severity depending on the nature and realism of the simulated environment and the methodologies of the training process~\cite{distribution}. This distribution gap can effect accuracy and real-world performance of a model greatly when feature representations extracted at runtime end up out of distribution from any learned representation, which result in misclassifications and a decrease in detector performance.

\section{Domain Adaptation for Object Detection}
Domain adaptation is a recently popular sub-field of machine learning that focuses on minimizing the data distribution difference between one set of labelled training data (source data) and another set of test-time inference data (target data). The expectation is that the target data looks vastly different (i.e. out of distribution) to the source data while sharing the same class instances. This degrades model performance as feature representations and decision boundaries learned on the source aren't directly applicable to the target. This can be attributed to source data bias learned during training, as the model is skewed towards the only encountered distribution of data. Therefore, domain adaptation techniques strive to minimize this source data bias and subsequent distribution shift by attempting to \textit{align} the source and target data distributions themselves. Recent methods achieve this through an unsupervised manner, where unlabelled target imagery is introduced to the model during training time. Domain classifiers (discriminators) are used throughout the model to attempt to classify which data domain feature representations come from. Adversarial training is employed, adding an additional training objective of trying to ``fool'' these discriminators (i.e. causing them to misclassify). Through this process robust features are learned that are domain agnostic, boosting recognition performance in the operational target domain while retaining class instance information.

\subsection{Domain Adaptation for Object Detection}
 As a relatively newer direction in computer vision research, many of the methods for domain adaptation have focused on the task of image classification with little work done in the realm of object detection. Domain Adaptive Faster R-CNN~\cite{domainrcnn} was one of the first architectures to bring domain adaptation methods to object detection by augmenting the Faster R-CNN two-stage object detection model. 
It focuses on aligning two key elements, image level representations and instance level representations. Image level representations refer to initial features that are extracted from the entire input image, while instance level representations handle those region proposals that are successfully localized and classified. Enforcing image level alignment ensures a more globally aligned feature distribution, while instance level alignment ensures that the model retains sufficient knowledge to classify objects in the target domain. 
 


Domain Adaptive YOLO (DA-YOLO)~\cite{dayolo} implements the same concept of image and instance level alignment as Domain Adaptive Faster R-CNN, but in the much faster object detection model of YOLOv3. Image level alignment in DA-YOLO is achieved through the use of three separate domain classifiers which operate on different outputs from the DarkNet-53 backbone feature extractor, each predicting domain labels of their respective feature map. This enforces strict alignment between local features while loosely aligning global ones. With no region proposals available, instance level alignment is enforced at each of YOLOv3's scale-handling layers, once again through three separate domain classifiers. Region of Interest (ROI) pooling is used to crop DarkNet feature maps based on detection outputs from each head before feeding into their respective domain classifiers. By obtaining the features that make up a given object detection before alignment, the use of ROI pooling aims to strictly align instances. This promotes greater invariance between the source and target domain's objects of interest themselves. 


\begin{figure*}
    \vspace{-10pt}
	\centering\includegraphics[width=0.75\textwidth]{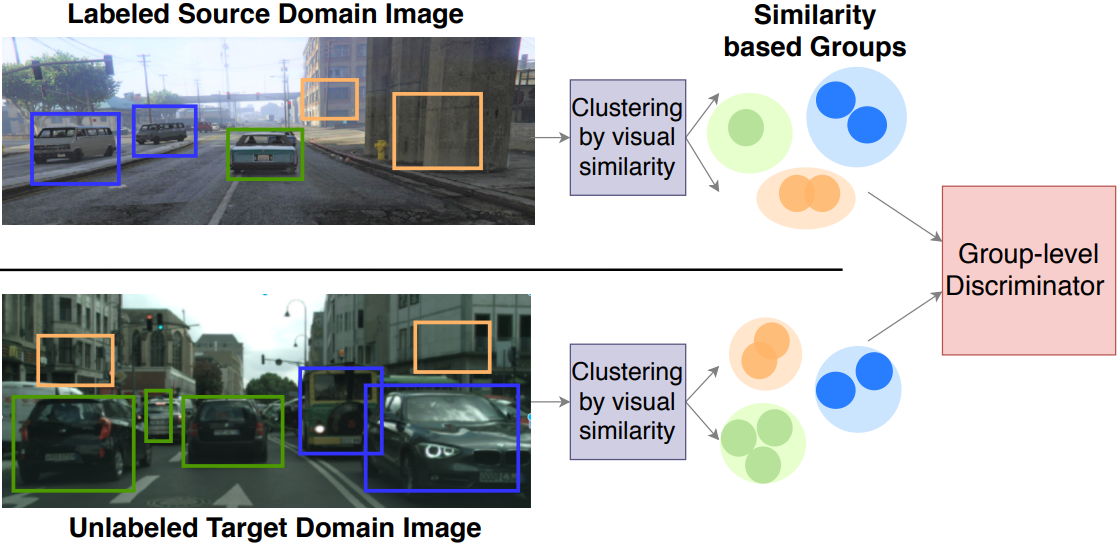}
	\caption{Clustering by visual similarity before inducing instance level alignment.}
	\label{fig:visga_groups}
	\vspace{-25pt}
\end{figure*}

\begin{figure*}[b]
    \vspace{-10pt}
	\centering
	\includegraphics[width=0.85\textwidth]{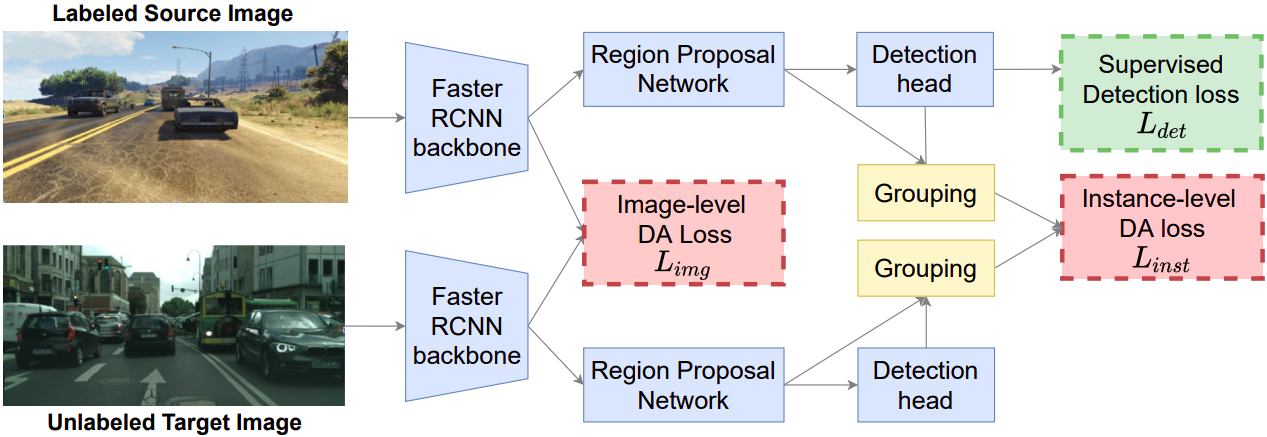}
	\caption{Domain adaptive object detection through visual similarity clustering in ViSGA.}
	\label{fig:visga}
	\vspace{-30pt}
\end{figure*}

Visually Similar Group Alignment (ViSGA)~\cite{visga} is another such work demonstrating domain adaptive object detection within the Faster R-CNN framework. ViSGA argues however that it is more beneficial to aggregate object proposals based on visual similarity before inducing instance level alignment, as shown in~\autoref{fig:visga_groups}. This simple tactic effectively aligns proposals from any spatial location in the image, and coarsely aligns the main feature clusters instead of attempting complete alignment of all object instances between the source and target domains, which heavily reduces complexity in the alignment process. The number of cluster groups is dynamically adjusted, ensuring the model retains enough capacity to represent inter-domain variance to each class. This similarity-based clustering approach, combined with adversarial training, allows ViSGA to achieve high invariance to source-target domain shift and be highly effective in simulation-to-real scenarios. The ViSGA architecture is shown in~\autoref{fig:visga}.

\section{You Only Crash Once}
\subsection{Methodology}
We propose \textit{You Only Crash Once} (YOCO), an end-to-end method for hazardous terrain detection and classification based on the YOLOv3 architecture with compatibility for the SC-LEARN embedded spacecraft co-processor, and augmented with visual similarity-based grouping and adversarial training for simulation-to-real domain adaptation. 
Opposed to DA-YOLO, we adopt visual similarity-based clustering for instance level alignment. The motivation for this adoption is based on three main factors:

\noindent {\bf Exploiting class variance}:  Clustering all object proposals by visual similarity before inducing instance level alignment allows the model to learn a more generalized representation for each type of terrain, which is crucial to obtain a broad range of detections during a spacecraft planetary landing scenario. With a many-vs-many approach to instance alignment, this technique ensures that all possible class features regardless of their location in the image space contribute to the alignment process. These clusters also represent and exploit a natural phenomenon when looking at the visual similarity between types of terrain, in which there can be very high intra and inter-class variance. Terrain of the same class may vary greatly in appearance amongst each other but be sufficiently dissimilar towards other classes (e.g. the visual appearance between a crater and another crater compared to a crater and a sand dune), a property easily modelled through clustering. With the assumption that terrain displays enough dissimilarity between classes, clustering by visual similarity is able to sufficiently group terrain instances of the same class (easily overcoming intra-class association issues) while forming large enough decision boundaries between classes. This leads to the formation of well-structured terrain groups, which allow for an efficient alignment process while simultaneously boosting recognition performance.
    
    
    
\noindent {\bf Improved recognition}:  The types of environments encountered during planetary landings can be incredibly feature-sparse. Planetary environments such as Mars exhibit much less edge and intensity gradients than common types of terrestrial environments (like cities), especially at high altitudes. This can pose a great challenge for the detector as the activation differences between feature maps can be extremely minimal. Clustering by visual similarity will therefore capture all such activations that lead to a successful detection of an object instance. This effectively \textit{attends} to those parts of the environment in which terrain hazards are found. By collecting and clustering these areas, the model is not only learning a representation that is robust across domains but learning a representation which is more distinguishable against the environment as well. 
    
\noindent {\bf Efficient alignment}: Clusters significantly reduce the challenge and complexity of alignment compared to modelling full data distributions. Estimating the prior distributions from which source and target data draw is a very difficult problem, one which many domain adaptation methods attempt to solve directly. The elegance of clustering before instance level alignment simplifies this potentially near-impossible problem, and has proven to be quite effective in modelling a generalized source-target distribution. 
    
    

\subsection{Implementation}

\subsubsection{\underline{Instance Clustering by Visual Similarity -}} 
With the absence of a region proposal network as in Faster R-CNN, sufficient engineering must be done to decide at which level to induce the grouping of feature maps before enforcing instance level alignment. There are many places within the YOLOv3 architecture where this can be accomplished, including at the output of the backbone feature extractor, amid the scale handling layers, or an aggregation of multiple feature maps throughout the model. In order to ensure that the entire network receives the effects of domain adaptation, we perform feature map clustering and subsequent instance domain alignment at each scale detection head. 
By enacting visual similarity-based clustering at each scale, we're ensuring that feature comparisons stay bounded within their local group. This ensures that activations produced by feature maps at larger scales don't oversaturate lower-scaled ones during alignment.
As detection heads predict the existence of an object through the activations in the final set of feature maps, the procedure to gather all feature maps that supply a given detection enables the production of a community-level representation consisting of all features that make up that object.

Given the sets of feature maps output from the final layer of each scale detection head, we form visual similarity-based groups through hierarchical agglomerative clustering. We apply agglomerative clustering to each set of feature maps at each scale detection head individually, resulting in three distinct clustering operations. For a given set of feature maps and beginning in a bottom-up fashion, each feature map is considered to be an individual cluster. Then, at each step, the two closest clusters are merged together according to a given distance metric. Following the ViSGA formulation of visual similarity-based clustering, we use cosine distance as our merging metric:
\begin{equation}
    dist(z_i, z_j) = 1 - \frac{z_i \cdot z_j}{||z_i||\text{ }||z_j||}
\end{equation}
where $z_i$ and $z_j$ show the $i$-th and $j$-th feature map respectively. With the goal to enhance performance in feature-sparse environments, we determine a set number of clusters prior to the clustering operation. Intuitively we seek to \textit{extract} prominent terrain features away from the more generally sparse background. This is unlike ViSGA, which sought to adaptively determine the number of cluster centers at clustering time. By enforcing a set number of cluster centers from the start, we aim to have the agglomerative clustering procedure attribute foreground terrain features of certain class instances together while excluding sparse background features. 
With each feature map exhibiting specific activations, the constraint on the number of clusters to find allows for all such activations to be associated together more easily. As a hyperparameter in the system, we empirically set the number of clusters to $N+1$, where $N$ is the number of classes. The agglomerative clustering procedure thus seeks to minimize a linkage function between members within each cluster. Also unlike ViSGA, we choose average linkage as our linkage criterion, defined by:
\begin{equation}
    \text{AvgLink}(A,B) = \frac{1}{N_A + N_B} \sum_{i=1}^{N_A} \sum_{j=1}^{N_B} \{dist(a,b): a \in A, b \in B \}
\end{equation}
where $A$ and $B$ are the sets of feature maps residing in two distinct clusters and $dist$ is our cosine distance merging metric. Following the convergence of the clustering procedure and the ViSGA formulation once again, we take the mean across feature maps in each cluster to construct a group representative embedding $Z_{c_i}$:
\begin{equation}
    Z_{c_i} = \frac{\sum_{i=0}^{N_{c_i}} z_i}{N_{c_i}}
\end{equation}
where $N_{c_i}$ is the number of instances assigned to the cluster $c_i$. Each group representative $Z_{c_i}$ is then fed to it's respective scale-level instance domain discriminator.

\subsubsection{\underline{Domain Classification -}}
We choose to apply domain adaptation through the use of four separate domain classifiers throughout our model. Each domain classifier seeks to simply predict whether the input features are from the source domain or the target domain. One domain classifier focuses on high-level image feature alignment (the image domain discriminator), and the remaining three are for low-level object feature alignment (instance domain discriminators), one for each scale detection head. Each discriminator is made up of the same architecture, which consists of a small number of CNN and dropout layers as well as a fully connected layer to predict the domain label. As the model improves it's supervised detections it also now aims to fool each discriminator by causing it to misclassify. Successfully fooling each discriminator means that the feature representations being learned by the model are sufficient enough to blur the distinction between source and target domains, heavily reducing training data bias and the underlying data distribution shift. 

\subsubsection{\underline{Domain Adaptive Network Training -}}
The overarching training objective of our network is the combined minimization over \textit{five} distinct losses: the supervised YOLO detection loss and each of the four domain classification losses. The supervised YOLO detection loss is comprised of localization loss, classification loss, and confidence loss. These three components strive to optimize the predictions from each grid cell, maximizing the models performance to detect, localize, and classify object instances. Grid cells that don't contain an object produce a confidence score of 0, which tends to overwhelm loss gradients of cells which actually do contain an object. To alleviate this the authors introduce two coefficients $\lambda_{coord}$ and $\lambda_{noobj}$ which reflect importance on the weight of detection loss and no object loss respectively. With this in mind the supervised YOLO detection loss $\mathcal{L}_{det}$ is defined as:
\vspace{-10pt}
\begin{multline}
    \mathcal{L}_{det} = \lambda_{coord} \sum^{S^2}_{i=0}  \sum^{B}_{j=0} \mathbbm{1}^{obj}_{ij} [(x_i - \hat{x}_i)^2 + (y_i - \hat{y}_i)^2] \\
    + \lambda_{coord} \sum^{S^2}_{i=0}  \sum^{B}_{j=0} \mathbbm{1}^{obj}_{ij} [(\sqrt{w_i} - \sqrt{\hat{w}_i})^2 + (\sqrt{h_i} - \sqrt{\hat{h}_i})^2] \\
    + \sum^{S^2}_{i=0}  \sum^{B}_{j=0} \mathbbm{1}^{obj}_{ij} (C_i - \hat{C}_i)^2 \\
    + \lambda_{noobj} \sum^{S^2}_{i=0}  \sum^{B}_{j=0} \mathbbm{1}^{noobj}_{ij} (C_i - \hat{C}_i)^2 \\
    + \sum^{S^2}_{i=0} \mathbbm{1}^{obj}_{i} \sum_{c \in \text{classes}} (p_i(c) - \hat{p}_i (c))^2
\label{eq:ldet}
\end{multline}
Interested readers are referred to the YOLO paper~\cite{yolo} for more information.

To successfully apply domain adaptation and guide the model towards learning domain invariant feature representations, the training objective seeks to cause each discriminator to wrongly predict which domain the features come from. This approach does not take any class information into account and instead focuses on the feature representations themselves. The features $F_d$ of domain $d$ ($d = 0$ for source and $d = 1$ for target) are fed to the discriminator $D$ which predicts the domain of the extracted features. The discriminator is trained by minimizing the binary cross-entropy loss defined as:
\begin{equation}
    \mathcal{L}_{disc} = -d\log(D(F_d)) - (1-d) \log(1 - D(F_d))
    \label{eq:disc}
\end{equation}
As we aim to have the features extracted by the model be indistinguishable by each discriminator, the domain adaptive training objective is to maximize the loss in~\autoref{eq:disc} with respect to the features $F_d$. This is achieved through adversarial training where a gradient reversal layer (GRL) is incorporated between the discriminator and portion of the network it connects to. GRL layers flip the sign of gradients flowing backward from the domain classification loss, which inform the model whether it is improving or degrading with respect to the quality of domain agnostic features being extracted. This facilitates a min-max game between each discriminator and the rest of the model.

With the YOLO detection and discriminator losses defined, we can now define the complete YOCO training loss as:
\begin{equation}
    \mathcal{L} = \mathcal{L}_{det} + \lambda_1 \mathcal{L}_{img} + \lambda_2 \mathcal{L}_{inst1} + \lambda_3 \mathcal{L}_{inst2} + \lambda_4 \mathcal{L}_{inst3}
\end{equation}
where $\mathcal{L}_{det}$ is the supervised YOLO detection loss, $\mathcal{L}_{img} = \mathcal{L}_{disc}$ and is the image discriminator loss, and $\mathcal{L}_{inst1}$, $\mathcal{L}_{inst2}$, and $\mathcal{L}_{inst3} = \mathcal{L}_{disc}$ and represent each scale detection head's instance discriminator loss respectively. $\lambda_1$, $\lambda_2$, $\lambda_3$, and $\lambda_4$ are hyperparameters that control the impact of each domain adaptation component separately. Unlike DA-YOLO that also employs multiple discriminators, we choose to weight each of them individually as to systematically tune detection performance between global and local features as well as by scale. The ability to tune detector performance at a high level is important for spacecraft landing in feature-sparse planetary environments, especially at high and mid altitudes where the detections are most critical before the commitment to a landing area. By weighting the factor of each domain adaptation component one can supply more criticality to a certain type or scale of feature depending on the operational environment.

\begin{figure*}
    \vspace{-10pt}
	\centering
	\includegraphics[width=.8\linewidth]{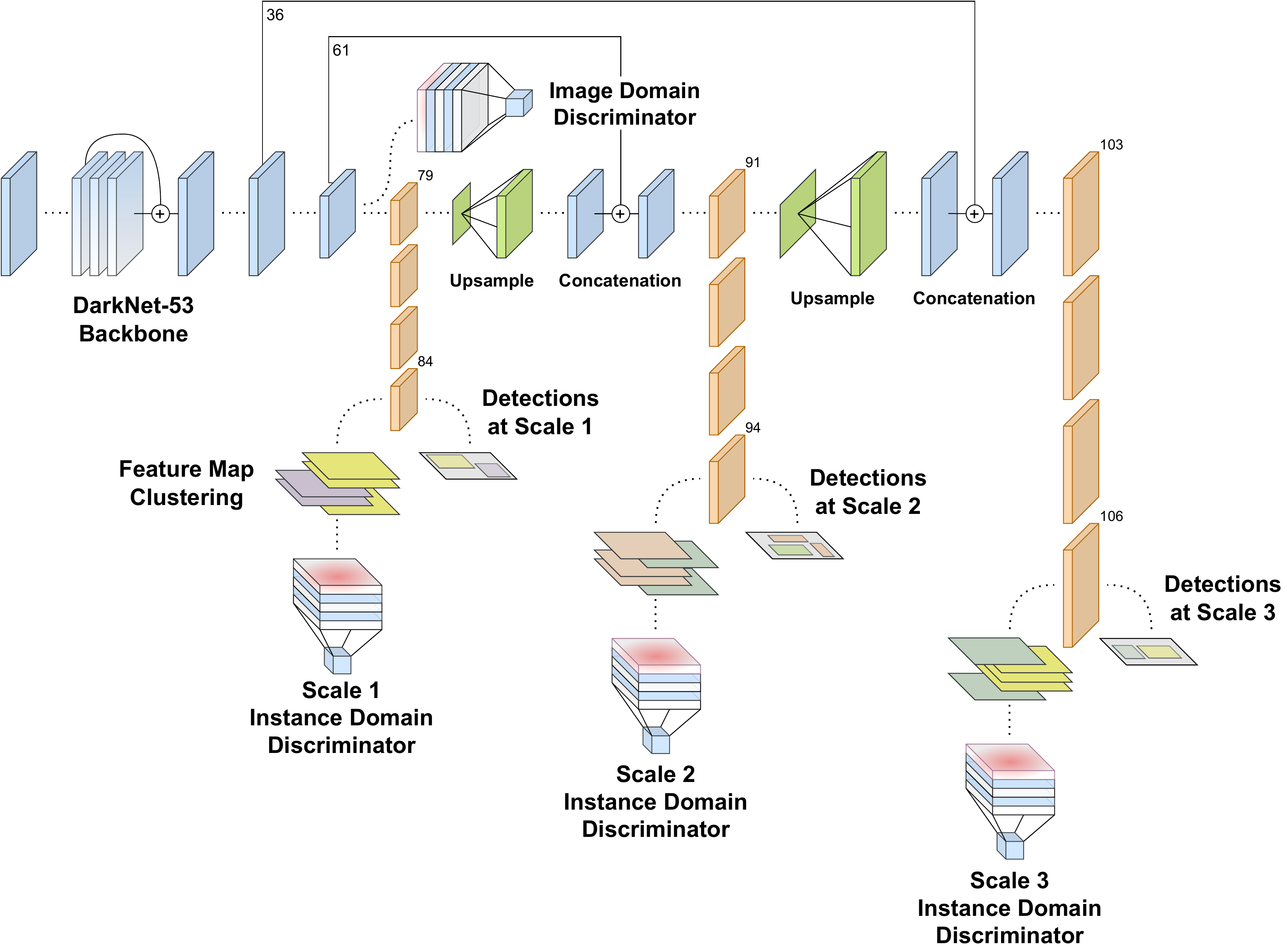}
	\caption{YOCO network architecture demonstrating image and instance domain alignment as well as feature map clustering by visual similarity.}
	\label{fig:yoco_arch} 
    \vspace{-25pt}
\end{figure*}

The use of visual similarity-based clustering as well as image and instance domain alignment through adversarial training make up YOCO, our improved YOLOv3 model for hazardous terrain detection and classification for autonomous spacecraft landing in feature-sparse planetary environments. The network architecture of YOCO is shown in~\autoref{fig:yoco_arch}.

\section{Experiments}
We perform various experiments in order to evaluate the performance of YOCO both quantitatively and qualitatively. These experiments occupy three categories:

\subsubsection{\underline{Sim-to-Real City -}}
With no labelled ground truth available for a planetary landing scenario, this experiment seeks to quantify YOCO performance using a standard terrestrial domain adaptation benchmark with the goal to detect cars in real urban environments with only simulated training data available. Although YOCO's design decisions are focused around feature-sparse planetary environments we still expect to see an increase performance over YOLOv3 in any simulation-to-real sceneario thanks to domain adaptive training.

\subsubsection{\underline{Sim-to-Sim Mars -}}
Before any analysis of hazardous terrain detection performance (even qualitative) can be done on real planetary imagery, labelled source data containing instances of representative terrain must be produced. This experiment sets up a more impactful sim-to-real landing scenario by developing a photo-realistic simulation of Mars to use as labelled source data. Quantitative metrics are reported on a testing set of the generated data to provide insights and analysis into what the model learns during training.

\subsubsection{\underline{Sim-to-Real Mars -}}
The final experiment is meant to qualitatively demonstrate YOCO's performance on real-world Mars images, both during past EDL operations and from orbital imagery. To demonstrate the effect of choosing representative target data, we show qualitative detections on \textit{two} datasets. The first contains image frames captured from the Mars Perseverance Rover's downward facing camera during it's landing sequence~\cite{mars2020edl}, while the second consists of terrain landmark imagery from the High Resolution Imaging Experiment (HiRISE) instrument onboard the Mars Reconnaissance Orbiter (MRO)~\cite{hirise}. These results showcase YOCO's ability to detect regions of hazardous terrain in the real-world on planetary environments, while also demonstrating robustness to even further out-of-distribution data by testing on Mars orbital imagery that was not introduced to the model during training.

\subsection{Experiment Setup}
All experiments follow the conventional unsupervised domain adaptation testing setup, in which full label annotations are provided for the source domain data and no labels are provided for the target domain. We train two models for each experiment, YOLOv3 with only source data available and YOCO with both source and target data available. A batch size of 32 with default anchor sizes are used, and each image is resized to 416x416 to fit the input of the YOLOv3 architecture. In order to conform to TensorFlow-Lite/Coral Edge TPU standard toolchains that enable model deployment on the SC-LEARN co-processor card, both YOLOv3 and YOCO are implemented in Tensorflow/Keras using an open source implementation as a baseline\footnote{https://github.com/qqwweee/keras-yolo3}. Both models are initialized with pretrained weights from COCO~\cite{coco}, a standard large-scale object detection, segmentation, and captioning dataset. Each model is trained for roughly 100 epochs amongst two stages to promote stability. The first stage keeps all layers frozen except each scale detection head, and lasts 50 epochs. The model is then unfrozen for the remaining 50 epochs as to train the entire model. We use the Adam optimizer with a learning rate of $1e^{-3}$ for the first stage and $1e^{-4}$ for the second. Early stopping and learning rate reduction are used in order to obtain the highest performing model. When reporting quantitative metrics we output all object predictions in COCO dataset format and use the standard evaluation procedure from the COCO API. For all such reporting a confidence threshold of 0.001 and maximum number of 100 for detections was used.
Metrics that report on object sizes are kept as the COCO evaluation defaults, which are $area < 32^2$ for small, $32^2 < area < 96^2$ for medium, and $area > 96^2$ for large.

\subsection{Sim-to-Real City}
\begin{figure*}
\vspace{-10pt}
\centering
\centering

\subfloat{\includegraphics[width=0.49\linewidth]{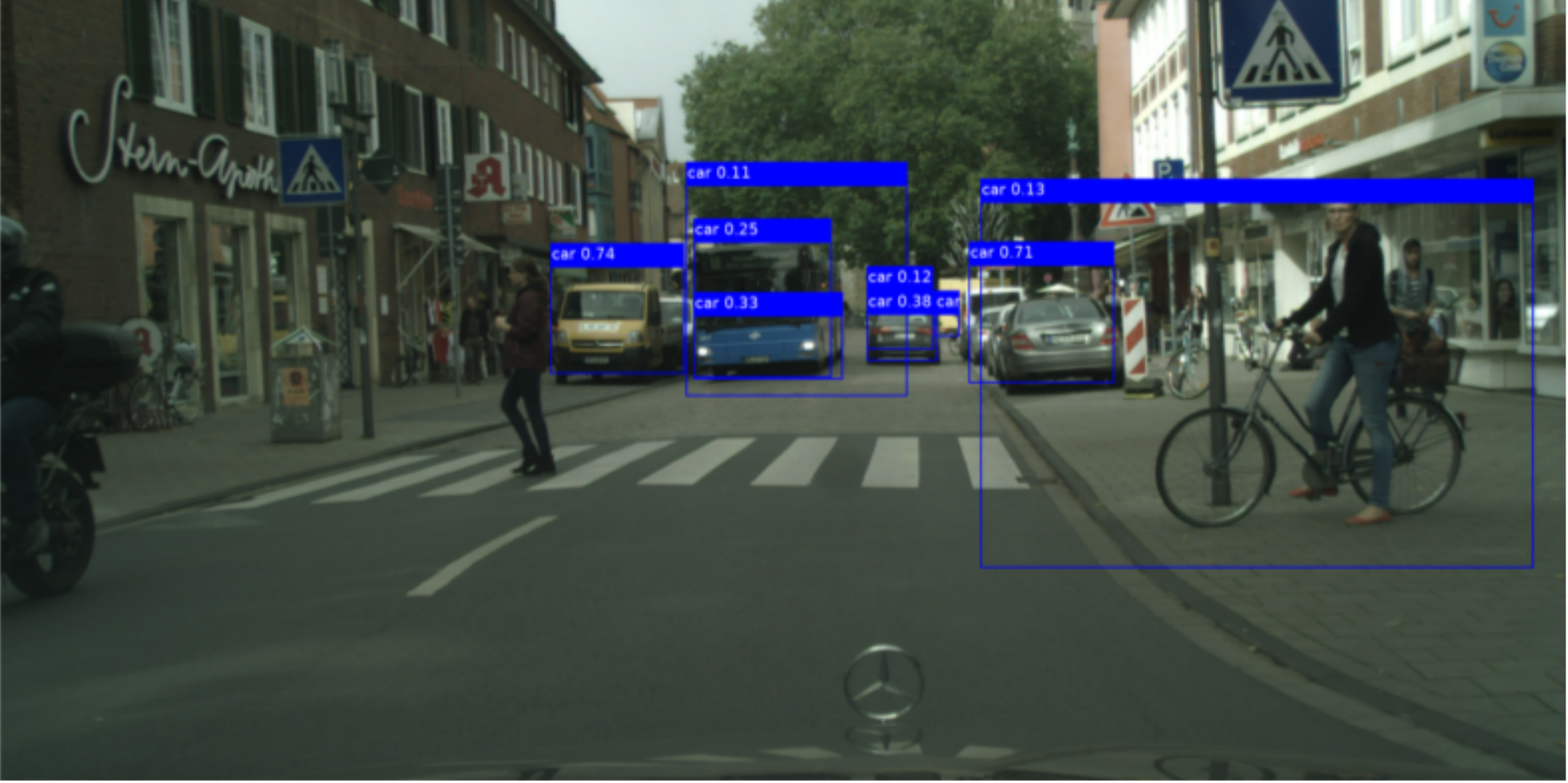}}
\hspace{1mm}%
\subfloat{\includegraphics[width=0.49\linewidth]{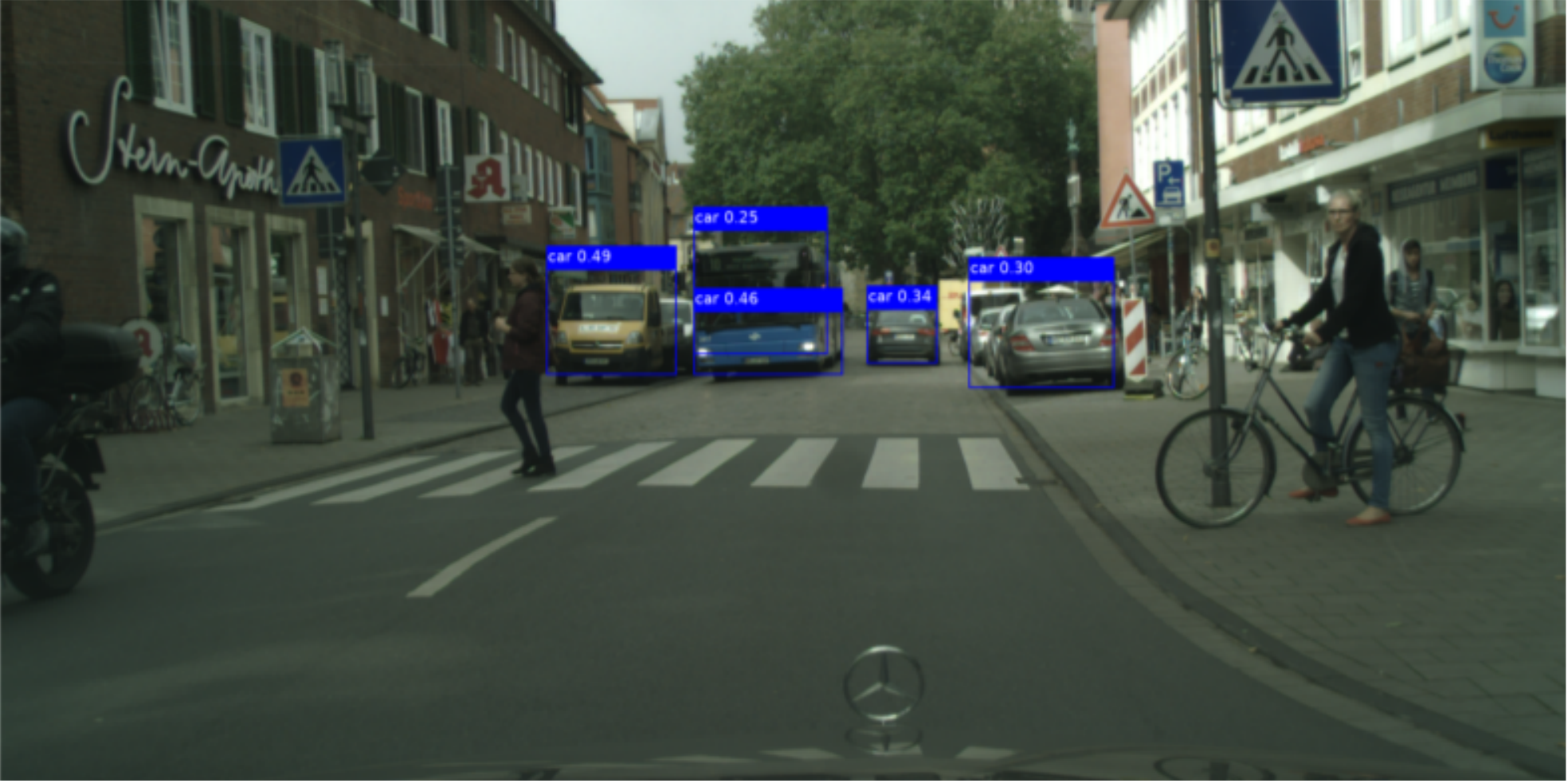}} \\
\vspace{-5pt}

\subfloat{\includegraphics[width=0.49\linewidth]{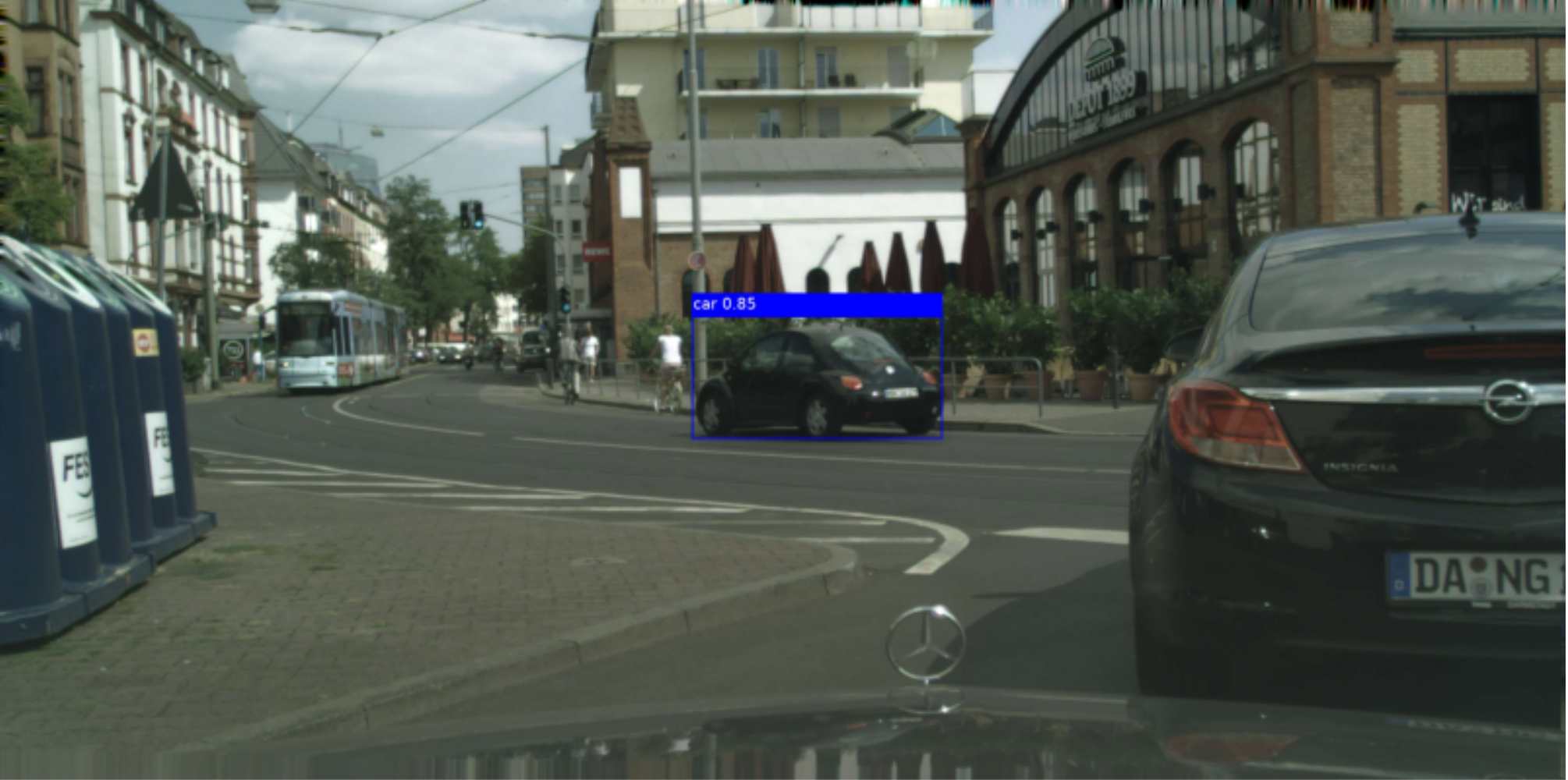}}
\hspace{1mm}%
\subfloat{\includegraphics[width=0.49\linewidth]{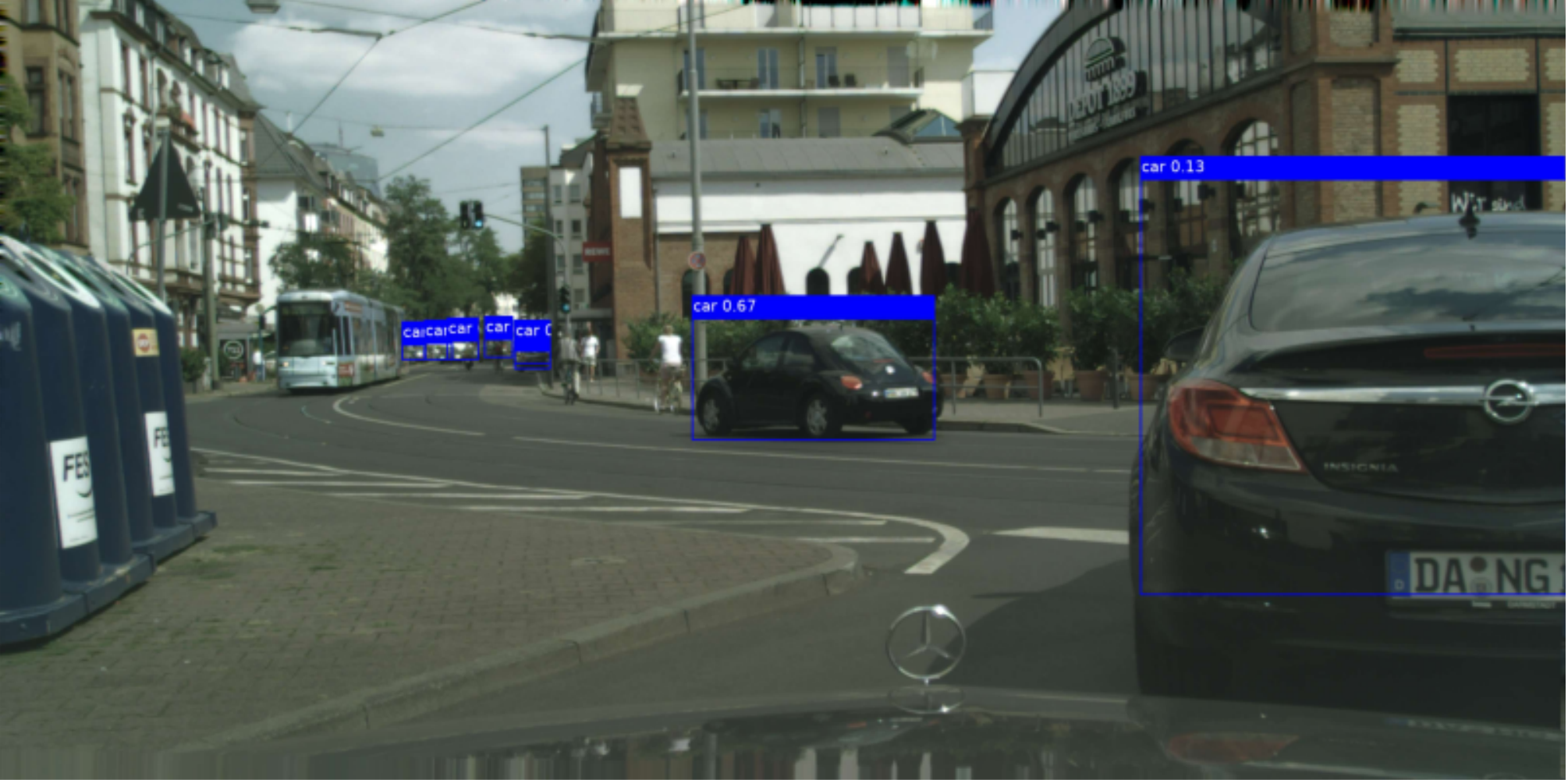}} \\
\vspace{-5pt}

\subfloat{\includegraphics[width=0.49\linewidth]{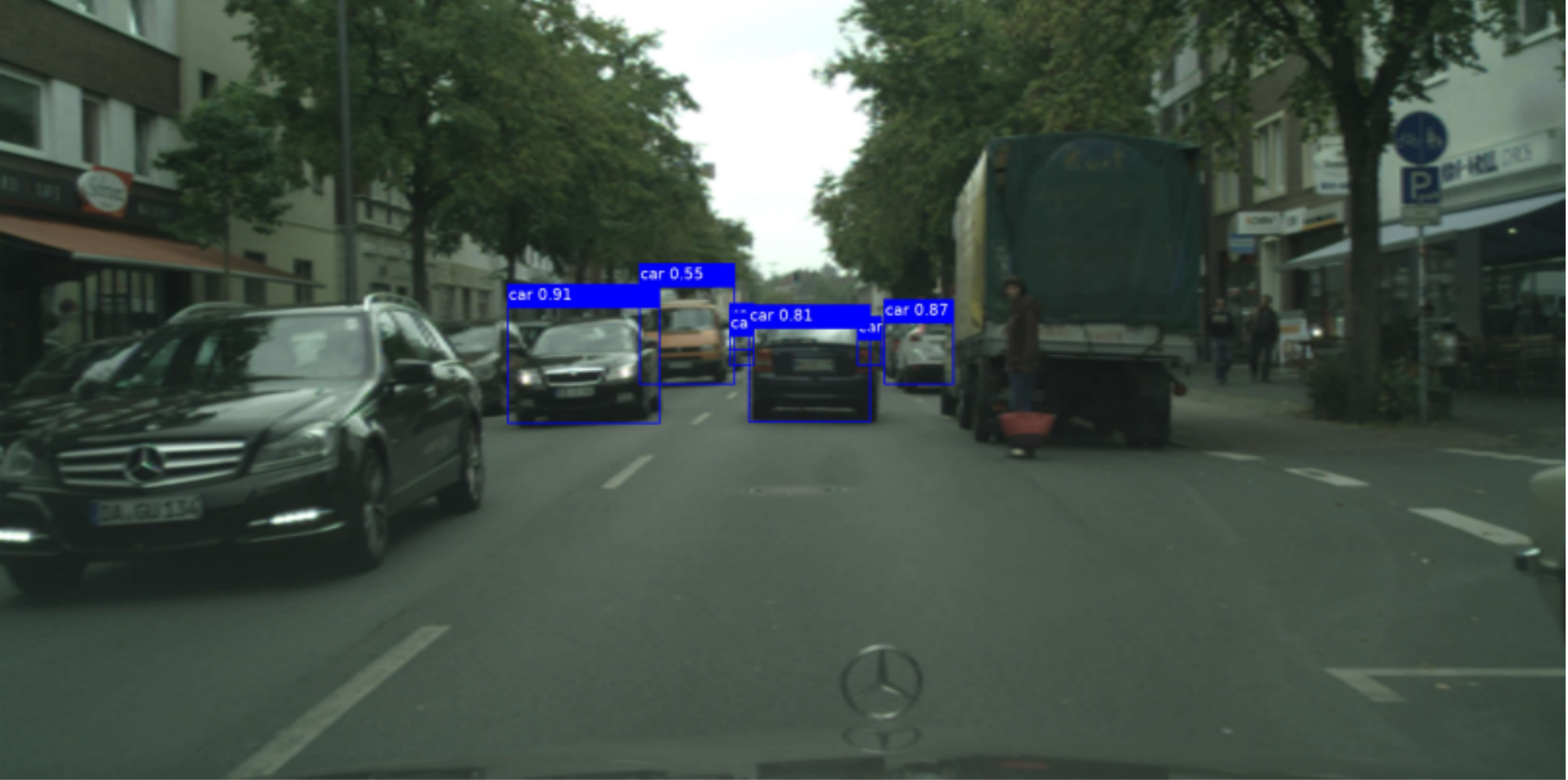}}
\hspace{1mm}%
\subfloat{\includegraphics[width=0.49\linewidth]{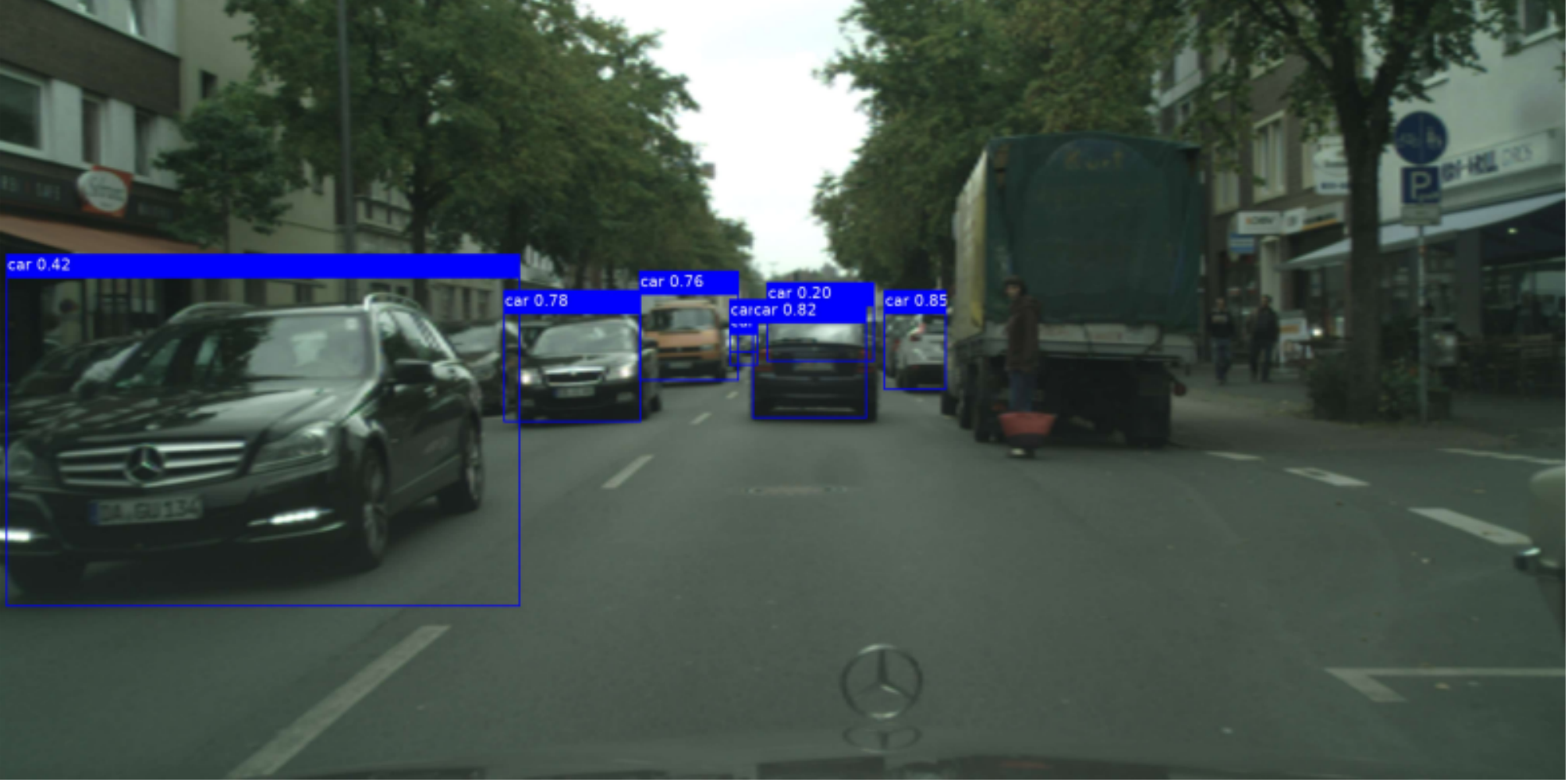}} \\
\vspace{-5pt}

\subfloat{\includegraphics[width=0.49\linewidth]{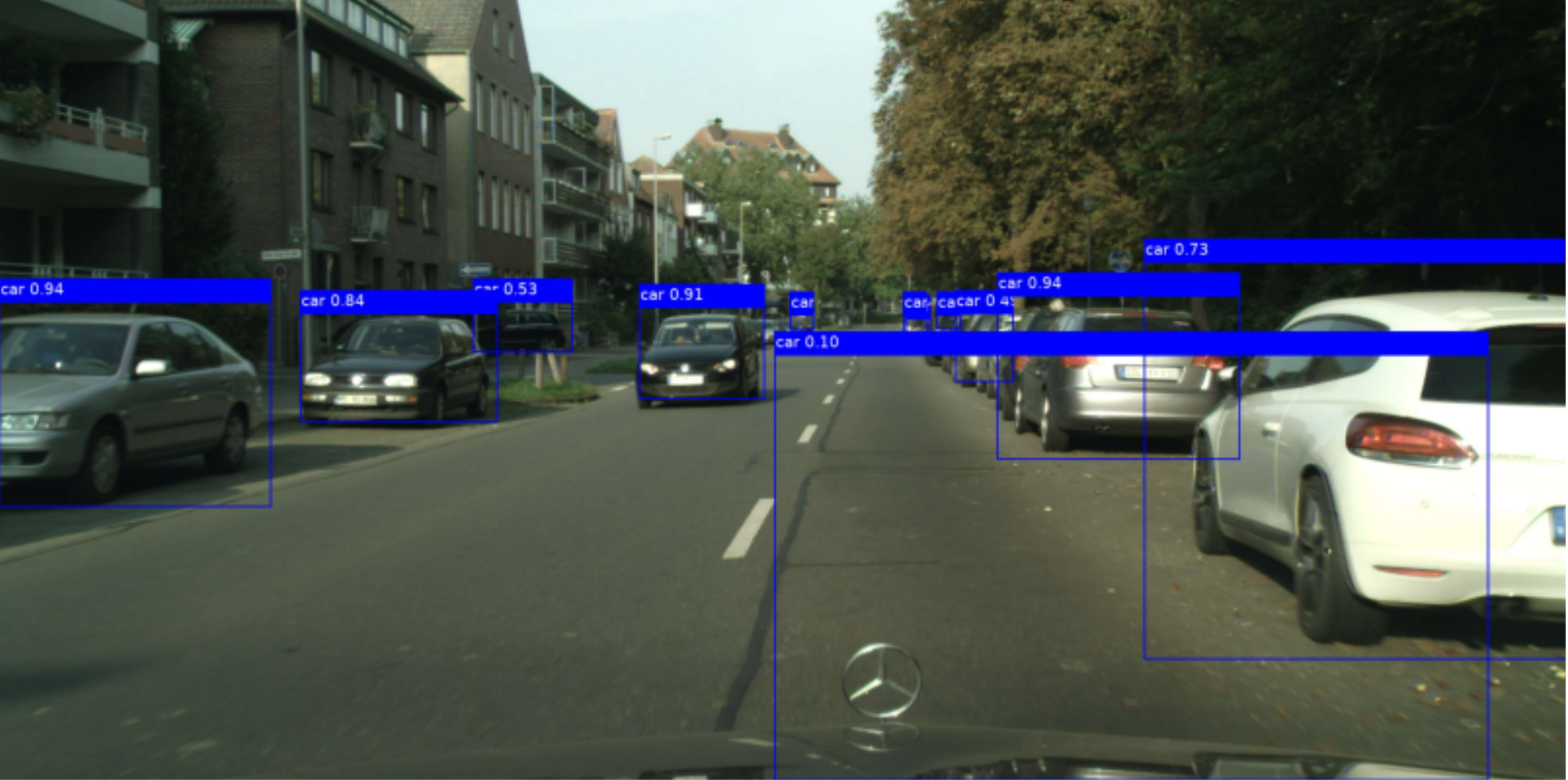}}
\hspace{1mm}%
\subfloat{\includegraphics[width=0.49\linewidth]{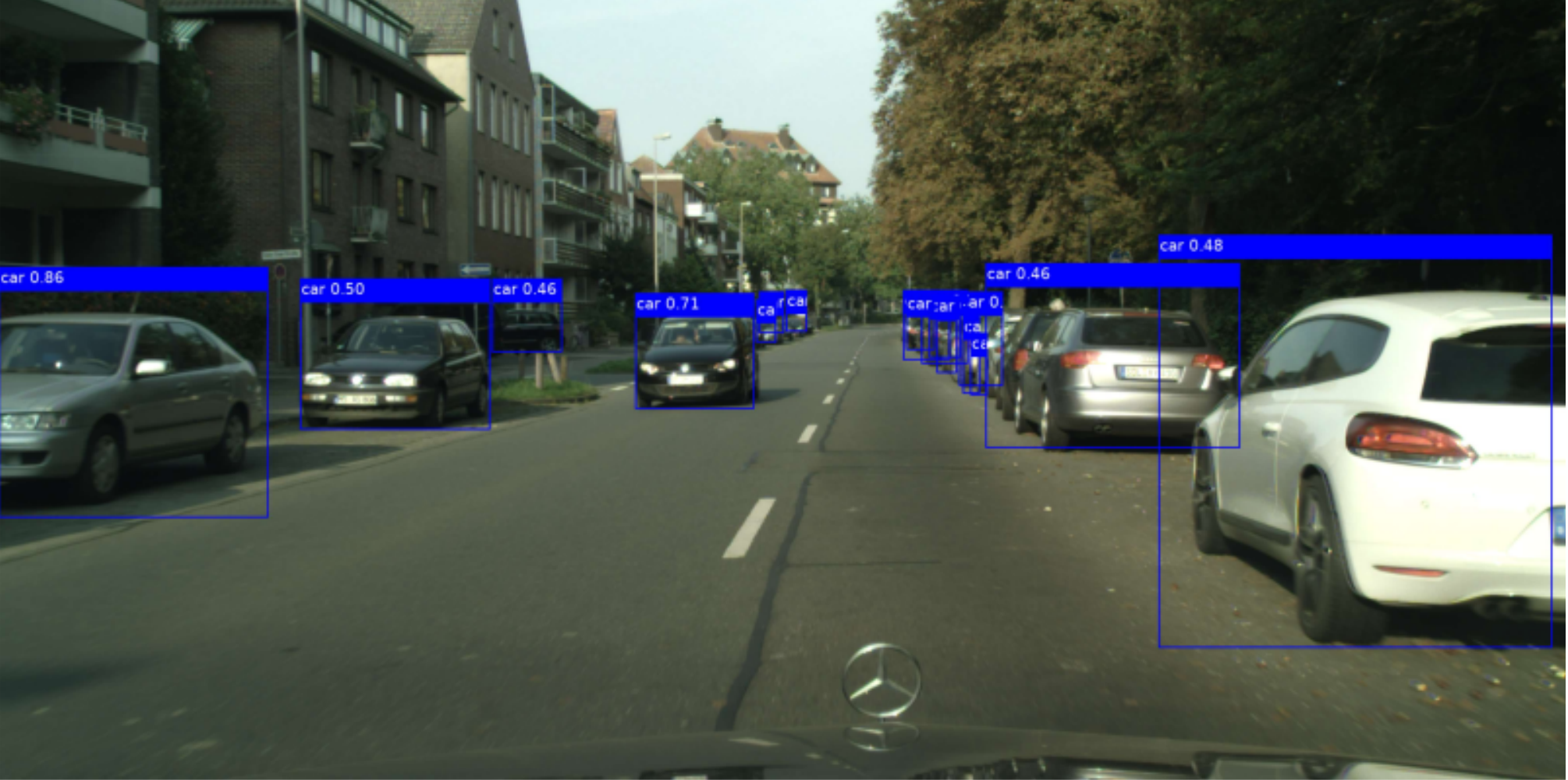}} \\
\vspace{-5pt}

\subfloat{\includegraphics[width=0.49\linewidth]{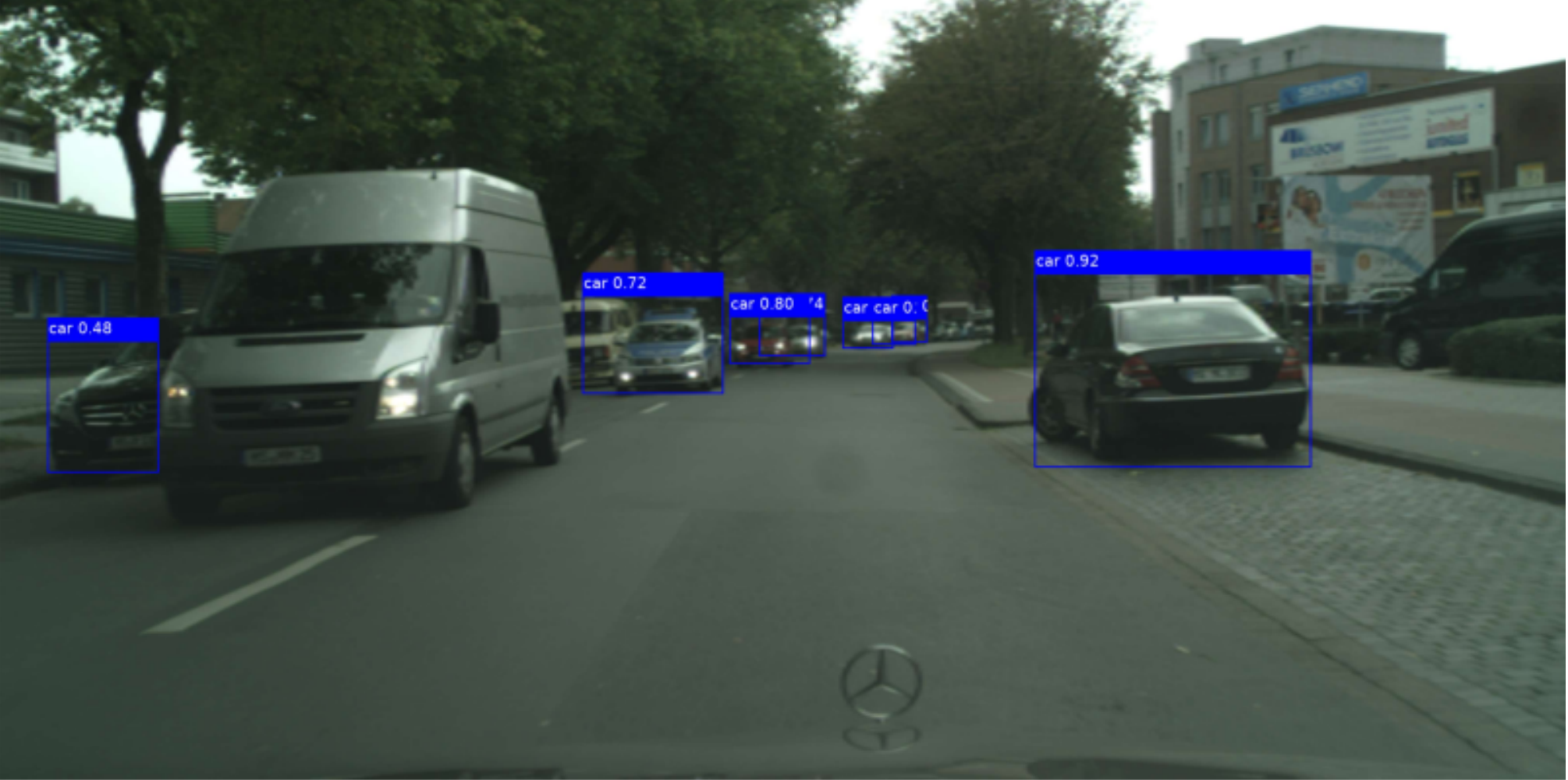}}
\hspace{1mm}%
\subfloat{\includegraphics[width=0.49\linewidth]{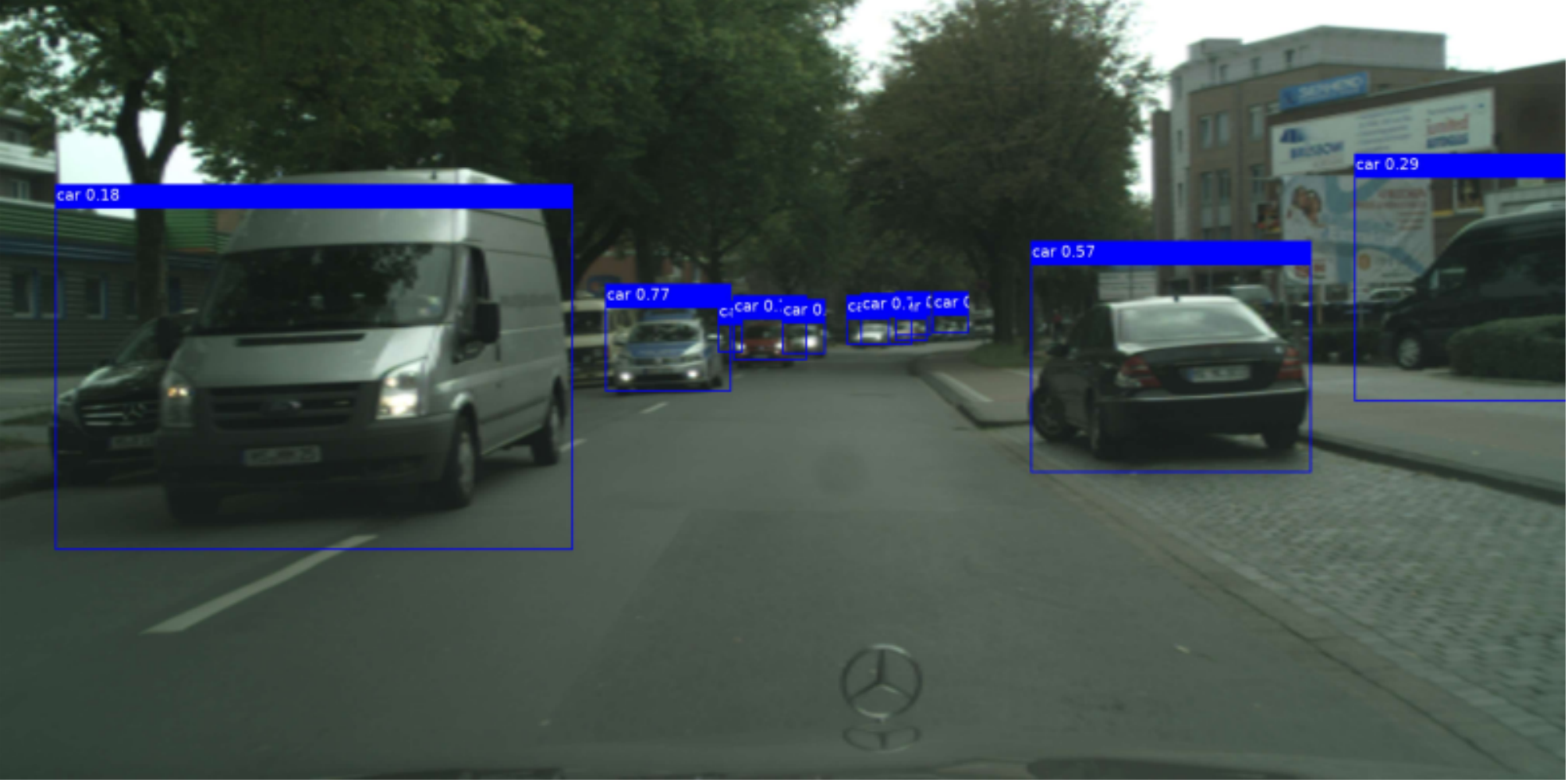}} \\

\caption{Qualitative examples from the Sim-to-Real City experiment between YOLOv3 detections (left) and YOCO detections (right). Detection counts of YOLOv3 vs YOCO from top to bottom: 9 vs 5, 1 vs 8, 7 vs 8, 12 vs 17, 8 vs 12.}
\label{fig:city_quals}
\vspace{-10pt}
\end{figure*}

These results aim to show the effectiveness of YOCO when trained through simulation while performing inference on real-world data through a standard terrestrial benchmark. SIM10k~\cite{sim10k} is used as labelled training data while testing is performed on Cityscapes~\cite{cityscapes}. SIM10k is a simulated dataset collected from the video game \textit{Grand Theft Auto V}, where ground truth bounding box labels are available for classes such as pedestrians and vehicles. Cityscapes exhibits the same classes as SIM10k but recorded in real urban environments across 50 European cities. For this experiment we focus on the \textit{car} class which is common procedure for this benchmark. For source data we use the entirety of the SIM10k dataset which contains 10,000 images. Cityscapes contains 5,000 images, from which we use 4,500 as YOCO target data and the remaining 500 as a test set for both systems. Average Precision (AP) and Average Recall (AR) metrics for the \textit{car} class are shown in~\autoref{tab:city_table}. Qualitative detection examples are shown in~\autoref{fig:city_quals}.

Our first observations is that YOCO outperforms YOLOv3 on the simulation-to-real scenario of SIM10k to Cityscapes on all metrics besides average recall on small sized objects, at which YOCO performs as well as YOLOv3. Average precision at 0.5 IoU shows a 3\% improvement in YOCO over YOLOv3. This is directly attributed towards the domain adaptive components in YOCO's design, which allow YOCO to learn more domain invariant features and adapt better to the real-world environment. A further observation is that YOCO outperforms YOLOv3 at detecting medium sized objects and \textbf{substantially} outperforms YOLOv3 at detecting large sized objects, where we see a 7.9\% and 42.9\% AP@IoU=0.5:0.95 increase respectively. We believe this drastic improvement to medium/large scale object detection has two contributing factors: the robust features learned through the introduction of target domain data and the clustering by visual similarity at the second and third scale detection heads. 

As medium and large scale objects consume a much larger fraction of the image, the ability to detect such objects is at the disposal of the quality of source data. With this experiment training on simulated data, the appearance of medium and large scale objects do not translate over to the real-world as well due to the resolution of the simulation. Compared to small objects, the discrepancy between visual fidelity from the simulation to real-world of larger objects will be much higher. YOLOv3 struggles to detect these objects due to this fidelity difference and directly demonstrates the effect of domain shift. In contrast YOCO \textbf{vastly} improves detection capability of medium and large scale objects despite this discrepancy, pointing to the model correctly learning robust features that greatly improve translation between the domains.

Further, we believe the factor of this domain discrepancy is made even smaller due to the procedure of visual similarity clustering at the medium and larger scale detection heads. Since medium and larger scale objects occupy more pixels in the input image we also expect activations to occupy more area in the feature maps at these detection heads. The clustering procedure on the feature maps that exhibit these activations will form well structured groups much more easily, as the computation of visual similarity between two such feature maps becomes an easier association problem. Producing the mean cluster representative from these well formed groups is then of higher quality, in which respective domain discriminators will make predictions on. By feeding the medium and large scale instance domain discriminators better representative embeddings, the alignment procedure has even greater effect as the classification accuracy on such embeddings will be high, triggering adversarial training and subsequently forcing the model to extract more agnostic feature representations.

Qualitative examples of detections between YOLOv3 and YOCO also demonstrate the ability of YOCO to not only better detect medium and larger scale objects through visual similarity clustering, but to improve detections at small scale and the overall quality of detections as well through image and instance domain adaptation. The first example (top row) demonstrates a YOLOv3 misclassification of the class \textit{car} to a cyclist, while also outputting many duplicate detections for the bus and small car down the road. YOCO does not make this misclassification and removes all but one of the duplicate detections, improving the overall quality of detections in this image. The next example highlights YOCO's robustness to domain shift as YOLOv3 only detects one vehicle compared to YOCO (correctly) detecting 8, mostly comprising small vehicles down the road as well as the larger vehicle in close proximity to the camera. This is demonstrated by the example in the third row as well, with the larger sized vehicle on the left of the frame being completely missed by YOLOv3 but detected by YOCO. A large misclassification from YOLOv3 can also be observed in the fourth example (second row from the bottom) at the large vehicle on the right hand side of the frame. Other vehicles parked on the right hand side of the road are also missed by YOLOv3. In contrast YOCO does not make this misclassification and also detects many of the vehicles on the right hand side even as they're bunched close together. Finally in the last example, we can see YOCO once again detecting two larger sized vehicles that YOLOv3 has missed (vans on the left and right side of the image) as well as a few smaller vehicles down the street. YOCO however misses detecting the smaller vehicle on the very left hand side of the frame, adjacent to the van, which YOLOv3 detects.

\begin{table*}
\vspace{-10pt}
\centering
\begin{tabular}{@{}cc|cc|cc@{}}
\toprule
\toprule
\multicolumn{2}{c|}{Class: car} & \multicolumn{2}{c|}{\textbf{YOLOv3}} & \multicolumn{2}{c}{\textbf{YOCO (Ours)}} \\ \midrule
@IoU     & Object Size   & AP & AR & AP & AR \\ \midrule
0.5:0.95 & all    & 13.3   & 30.2 & 14.7 & 31.9   \\
0.50     & all    & 30.1   & -    & 33.1 & -      \\
0.75     & all    & 10.0   & -    & 11.7 & -      \\
0.5:0.95 & small  & 9.4    & 23.8 & 9.6  & 23.8   \\
0.5:0.95 & medium & 34.2   & 61.6 & 42.1 & 67.6   \\
0.5:0.95 & large  & 12.0   & 27.0 & 54.9 & 77.2   \\ \bottomrule
\end{tabular}
\caption{Average Precision (AP) and Average Recall (AR) metrics between YOLOv3 and YOCO for the \textit{car} class on the Cityscapes test set, computed against various IoU thresholds and object sizes.}
\label{tab:city_table}
\vspace{-20pt}
\end{table*}

\subsection{Sim-to-Sim Mars}
In preparation for simulation-to-real experiments with true planetary landing and orbital images from Mars, this experiment serves to generate representative photo-realistic images of Martian terrain for required annotated source data while simultaneously enabling an ablation study over the types of terrain detections YOLOv3 and YOCO make from a purely simulation standpoint. We create a photo-realistic simulation of Martian terrain in Blender~\cite{blender}, a widely used and open source software for 3D modelling and animation. We generate 8,851 images from our simulation containing \textit{crater}, \textit{sand dune}, and \textit{mountain} classes, with 121,069, 43,711, and 19,155 instances respectively. 7,966 images are used as source domain training data and 885 are withheld as a testing set. Also in preparation for simulation-to-real experiments on Mars images, we use 3,659 image frames from the Mars Perseverance Rover landing as unlabelled target data when training YOCO.

\subsubsection{\underline{Blender Simulation Development of Martian Terrain -}}
\begin{figure}[H]
\vspace{-10pt}
\centering
\subfloat{\includegraphics[width=.25\linewidth]{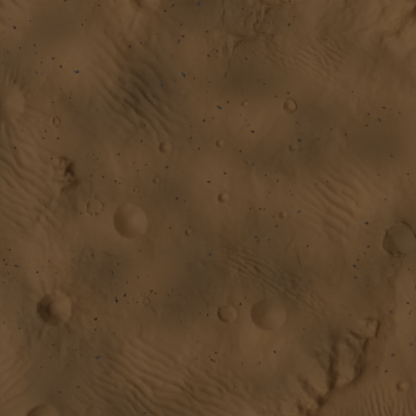}} 
\hspace{10pt}
\subfloat{\includegraphics[width=.25\linewidth]{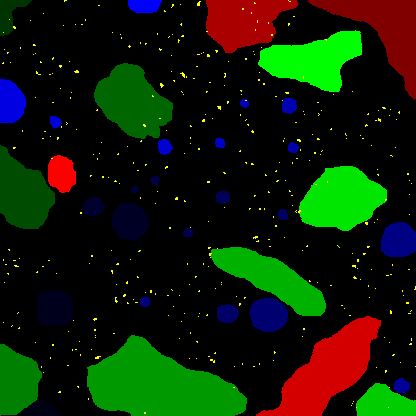}} 
\hspace{10pt}
\subfloat{\includegraphics[width=.25\linewidth]{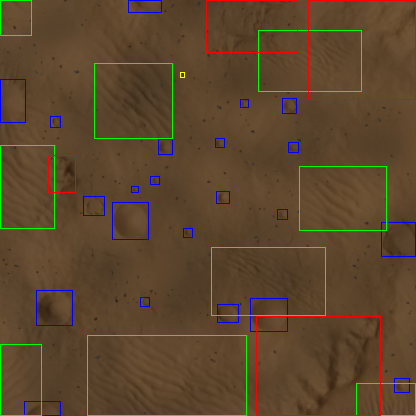}}\\
\subfloat{\includegraphics[width=.25\linewidth]{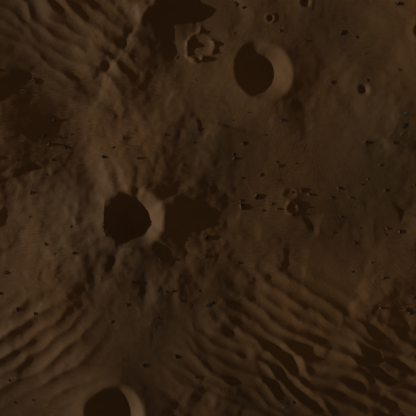}} 
\hspace{10pt}
\subfloat{\includegraphics[width=.25\linewidth]{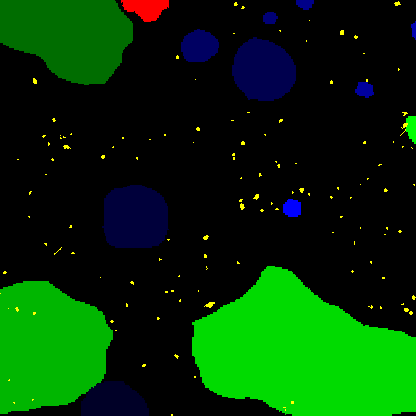}} 
\hspace{10pt}
\subfloat{\includegraphics[width=.25\linewidth]{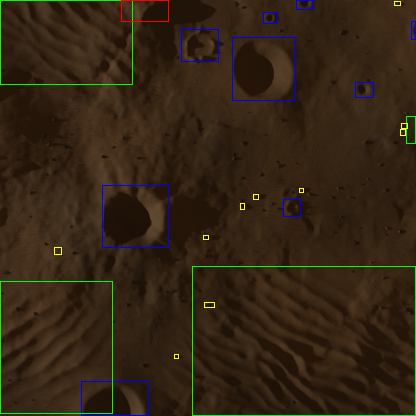}}\\
\caption{Two example images generated from the Blender simulation with their corresponding feature segmentation masks and bounding box output.}
\label{fig:mars_sim}
\vspace{-20pt}
\end{figure}

Blender was used to generate training data of the martian surface.  This was done by creating a large region of martian surface containing a range of the various feature classifications, and then splitting off each instance of a particular feature into its own geometry.  The camera position and illumination conditions were then randomly sampled and used to render an image.  Because each instance was its own individual geometry, a separate instance mask is generated for each image to show which pixels corresponded to which feature.  These instance segmentation masks are then be used to generate the bounding boxes used for training.  Figure \ref{fig:mars_sim} demonstrates this process for two different images.

~\autoref{tab:mars_sim_table_ap} and~\autoref{tab:mars_sim_table_ar} show the average precision and average recall at different IoU thresholds and object sizes for each class in the simulated dataset between YOLOv3 and YOCO. From both training and testing on the simulated Martian terrain, we observe similar patterns as with the Sim-to-Real City experiment. For both average precision and average recall on all classes, object detection performance is once again improved in YOCO compared to YOLOv3. Similarly, of particular interest is large sized objects, in which YOLOv3 significantly struggles to detect. Both average precision and average recall for large objects were 0\% in YOLOv3 for all but the \textit{sand} class. Conversely, YOCO is able to detect these instances which improves overall average precision and average recall for large objects by a very large margin. With large objects mostly dominated by the \textit{sand} class, this points to YOLOv3 struggling to discern large sand dune features against either smaller ``ripples'' in the terrain or from the background surface. We can attribute the performance increase in YOCO once again to the visual similarity clustering procedure. As mentioned previously, we assume the visual similarity of larger objects will be easier to cluster. With the clustering procedure also operating on source domain data as well as target domain data, YOCO is demonstrating the ability to more accurately capture the representations of large and more challenging terrain such as \textit{sand} and \textit{mountains} in the simulated data.

Further, we empirically observed lower supervised detection loss in YOLOv3 than in YOCO. With the aforementioned improvements to YOCO performance on the simulated testing set, we believe this behavior is attributed to a potential regularization towards overfitting within YOCO by the addition of target domain data. Although each system is trained and tested on simulated data (data from the same domain), the YOCO model is still effected by domain adaptation that the image and instance discriminators are forcing through adversarial training. With the inclusion of target data to the YOCO model, the domain adaptation components begin to ``pull'' feature representations away from source domain bias as training progresses. Although not definitively interesting by itself, this observation demonstrates the powerful effect that the four domain discriminators have on the model and shows that the domain adaptation alignment procedure is working correctly even with such a small amount of target domain data being introduced.

\begin{table*}
\vspace{-10pt}
\centering
\begin{tabular}{@{}cc|cccc|cccc@{}}
\toprule
\toprule
\multirow{2}{*}{@IoU} & \multirow{2}{*}{Object Size} & \multicolumn{4}{c|}{\textbf{YOLOv3}} & \multicolumn{4}{c}{\textbf{YOCO (Ours)}} \\
         &        & crater & sand & mountain & mAP  & crater & sand & mountain & mAP  \\ \midrule
0.5:0.95 & all    & 28.6   & 12.6 & 4.3      & 15.2 & 41.3   & 43.2 & 29.1     & 37.9 \\
0.50     & all    & 56.6   & 17.6 & 7.5      & 27.3 & 73.1   & 69.5 & 58.7     & 67.1 \\
0.75     & all    & 25.5   & 15.0 & 4.3      & 14.9 & 42.2   & 50.7 & 24.1     & 39.0 \\
0.5:0.95 & small  & 27.3   & 2.2  & 9.9      & 13.1 & 39.7   & 3.1  & 31.5     & 24.8 \\
0.5:0.95 & medium & 45.6   & 18.5 & 5.4      & 23.1 & 52.7   & 45.9 & 28.1     & 42.2 \\
0.5:0.95 & large  & 0.0    & 0.7  & 0.0      & 0.2  & 15.8   & 47.0 & 34.1     & 32.3 \\ \bottomrule
\end{tabular}
\caption{Average Precision (AP) between YOLOv3 and YOCO for \textit{crater}, \textit{sand}, and \textit{mountain} classes on the synthetic Mars terrain, computed against various IoU thresholds and object sizes.}
\label{tab:mars_sim_table_ap}
\vspace{-5pt}
\end{table*}

\begin{table*}
\centering
\vspace{-10pt}
\begin{tabular}{@{}cc|cccc|cccc@{}}
\toprule
\toprule
\multirow{2}{*}{@IoU} & \multirow{2}{*}{Object Size} & \multicolumn{4}{c|}{\textbf{YOLOv3}} & \multicolumn{4}{c}{\textbf{YOCO (Ours)}} \\
         &        & crater & sand & mountain & mAR  & crater & sand & mountain & mAR  \\ \midrule
0.5:0.95 & all    & 45.9   & 18.3 & 6.1      & 23.4 & 63.7   & 67.9 & 61.5     & 64.4 \\
0.5:0.95 & small  & 43.7   & 7.2  & 15.3     & 22.0 & 61.8   & 19.0 & 50.3     & 43.7 \\
0.5:0.95 & medium & 58.3   & 25.9 & 7.1      & 30.4 & 74.1   & 71.2 & 61.6     & 69.0 \\
0.5:0.95 & large  & 0.0    & 0.1  & 0.0      & 0.0  & 27.3   & 69.7 & 66.8     & 54.6 \\ \bottomrule
\end{tabular}
\caption{Average Recall (AR) between YOLOv3 and YOCO for \textit{crater}, \textit{sand}, and \textit{mountain} classes on the synthetic Mars terrain, computed against various IoU thresholds and object sizes.}
\label{tab:mars_sim_table_ar}
\vspace{-20pt}
\end{table*}

\subsection{Sim-to-Real Mars}
Continuing from the Sim-to-Sim Mars experiment, this set of experiments evaluates YOCO performance against YOLOv3 on real-world Mars images from two different scenarios: the landing of the Perseverance Rover and HiRISE landmark images from MRO. As no dataset contains ground truth bounding box labels, we aim to show the effectiveness of YOCO and it's design decisions towards feature-sparse planetary landing operations empirically. Both scenarios use 3,659 random images from the Mars Perseverance Rover landing sequence as unlabelled target domain data. Mars Perseverance landing images that are used for qualitative analysis are withheld from YOCO's target domain training data.

\subsubsection{\underline{Mars Perseverance EDL -}}

\begin{figure*}
\centering
\subfloat{\includegraphics[width=0.24\linewidth]{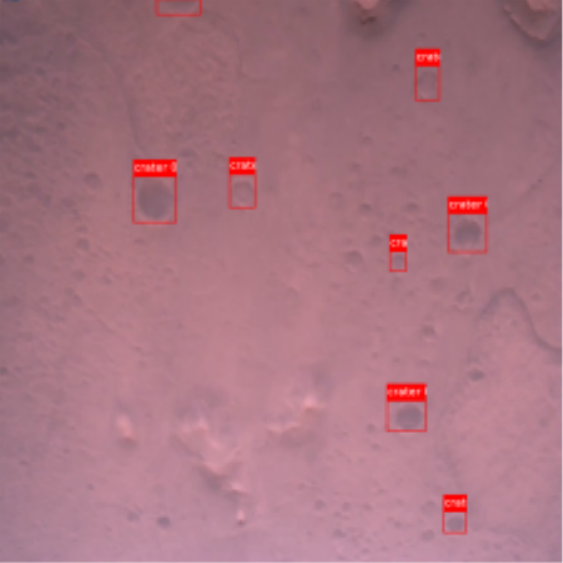}}
\hspace{1mm}%
\subfloat{\includegraphics[width=0.24\linewidth]{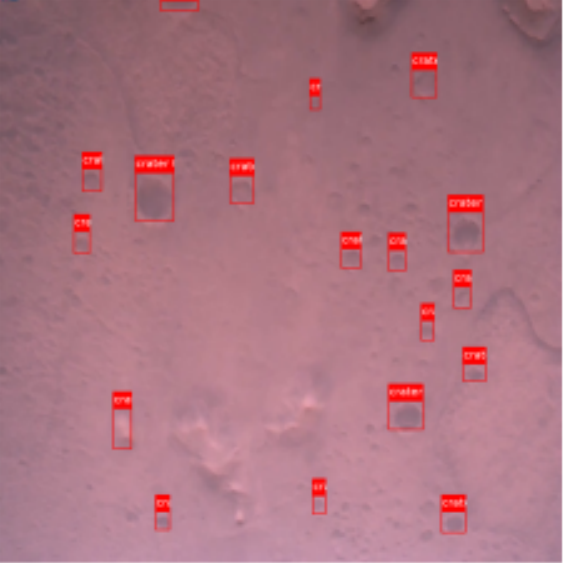}}
\hspace{5pt}%
\subfloat{\includegraphics[width=0.24\linewidth]{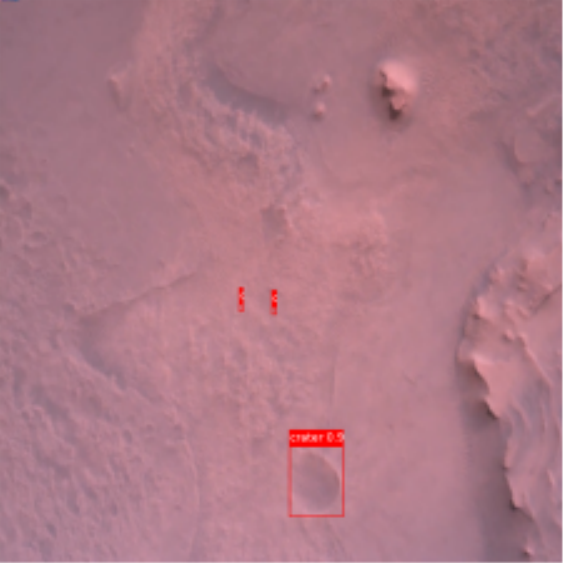}}
\hspace{1mm}%
\subfloat{\includegraphics[width=0.24\linewidth]{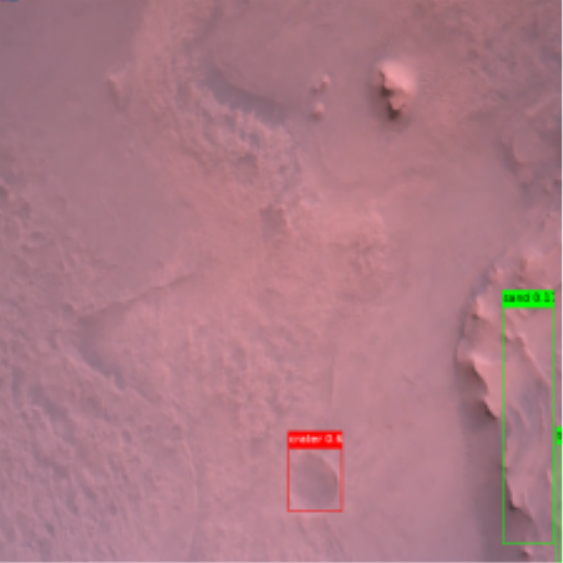}} \\
\vspace{-5pt}

\subfloat{\includegraphics[width=0.24\linewidth]{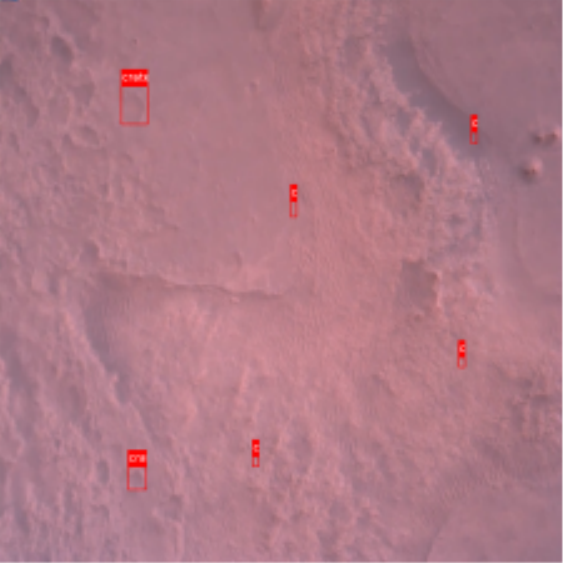}}
\hspace{1mm}%
\subfloat{\includegraphics[width=0.24\linewidth]{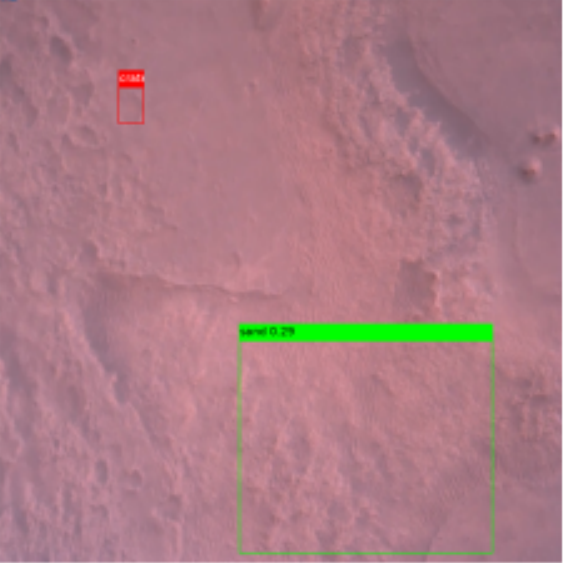}}
\hspace{5pt}%
\subfloat{\includegraphics[width=0.24\linewidth]{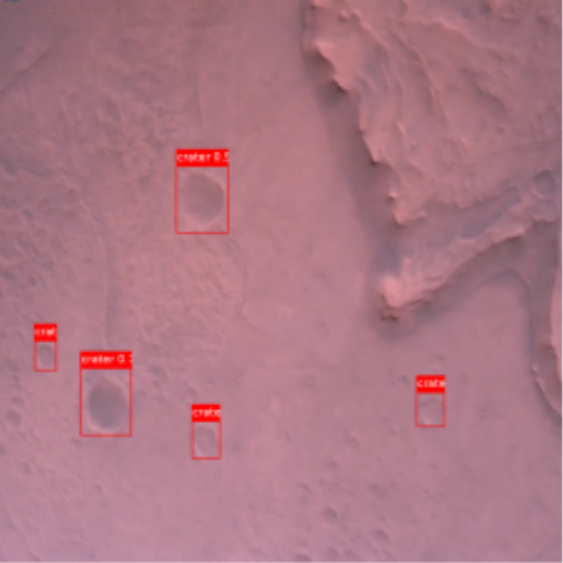}}
\hspace{1mm}%
\subfloat{\includegraphics[width=0.24\linewidth]{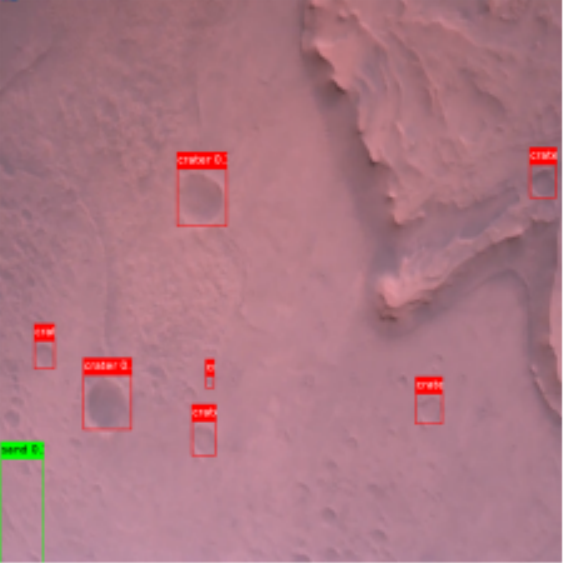}} \\

\caption{Qualitative examples from the Sim-to-Real Mars experiment between YOLOv3 (left) and YOCO (right) detections on Mars Perseverance EDL images. Red boxes represent crater detections, green represents sand, and blue represents mountain.}

\label{fig:m2020_quals}
\vspace{-30pt}
\end{figure*}

\begin{figure*}
\centering
\subfloat{\includegraphics[width=0.24\linewidth]{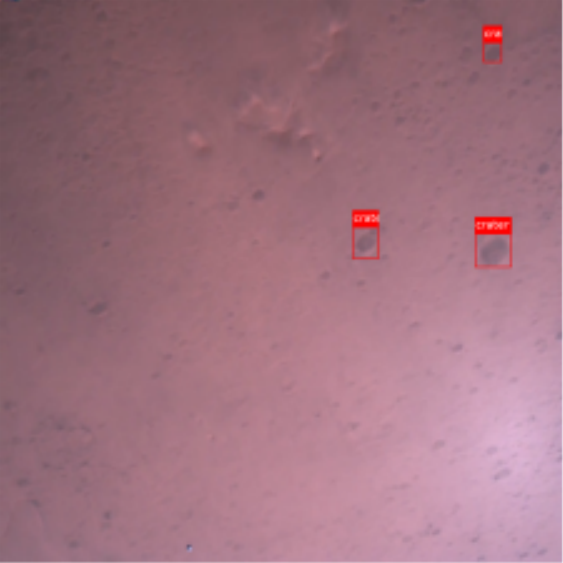}}
\hspace{1mm}%
\subfloat{\includegraphics[width=0.24\linewidth]{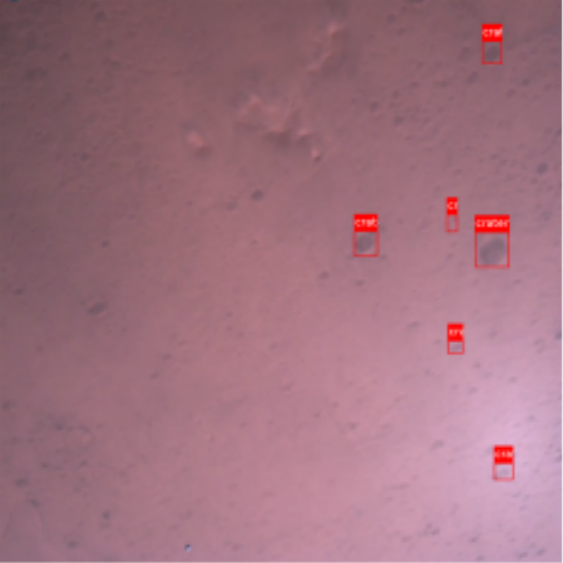}}
\hspace{5pt}%
\subfloat{\includegraphics[width=0.24\linewidth]{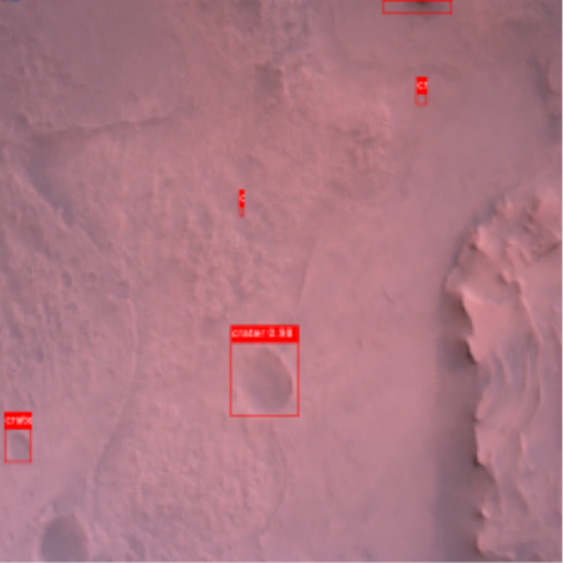}}
\hspace{1mm}%
\subfloat{\includegraphics[width=0.24\linewidth]{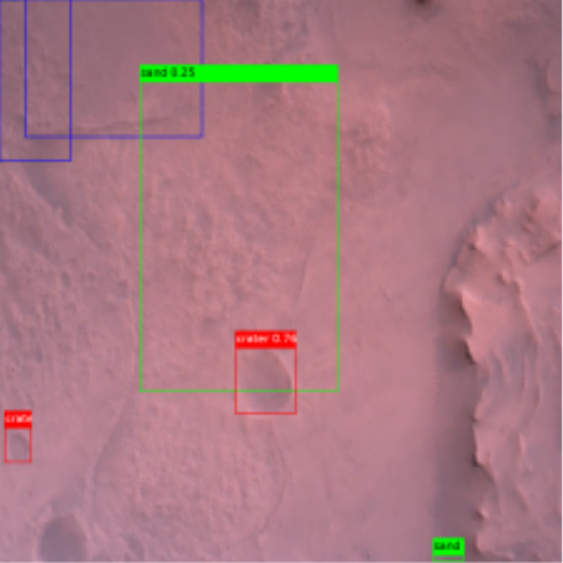}} \\
\vspace{-5pt}

\subfloat{\includegraphics[width=0.24\linewidth]{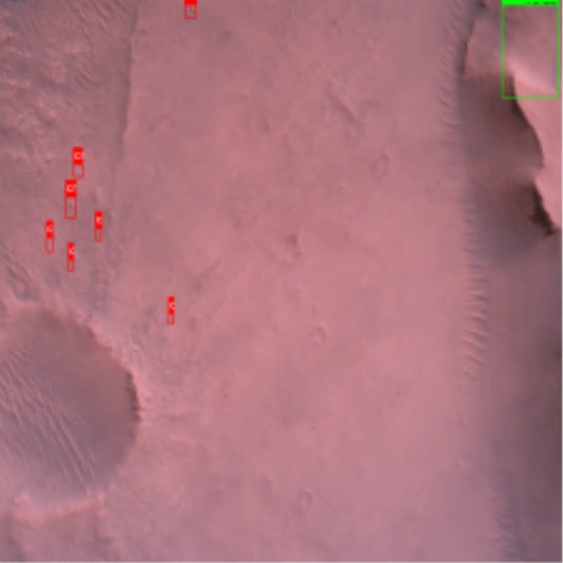}}
\hspace{1mm}%
\subfloat{\includegraphics[width=0.24\linewidth]{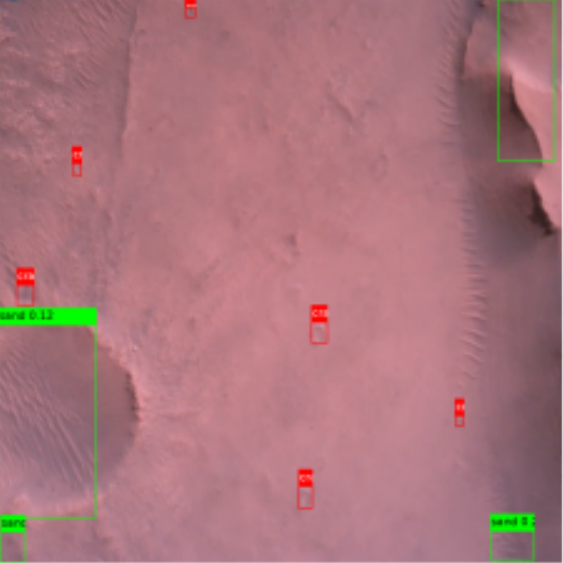}}
\hspace{5pt}%
\subfloat{\includegraphics[width=0.24\linewidth]{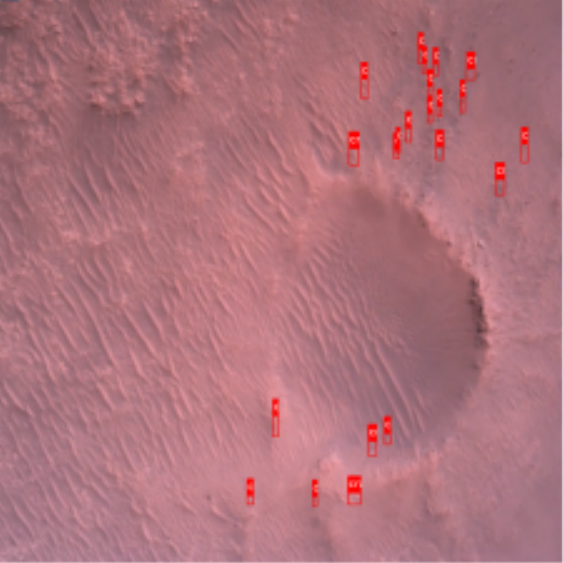}}
\hspace{1mm}%
\subfloat{\includegraphics[width=0.24\linewidth]{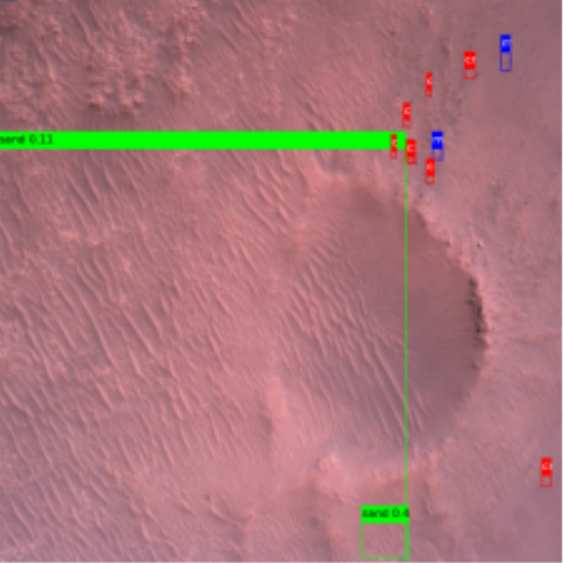}} \\

\caption{Additional qualitative examples from the Sim-to-Real Mars experiment between YOLOv3 (left) and YOCO (right) detections on Mars Perseverance EDL images.}

\label{fig:m2020_quals2}
\vspace{-20pt}
\end{figure*}

~\autoref{fig:m2020_quals} and~\autoref{fig:m2020_quals2} show qualitative detection examples between YOLOv3 and YOCO on images captured during the Mars Perseverance Rover's landing sequence. Similarly to the experiment of Sim-to-Real City, our first observation is that YOCO greatly improves upon the ability to accurately make detections in the real-world compared to YOLOv3. Not only is there a greater number of detections in some cases, but in every example the overall quality of the detections are improved. Many misclassifications from YOLOv3 no longer appear by YOCO, and terrain that YOLOv3 fails to detect at all are able to be detected by YOCO. This once again is attributed in part by domain adaptive training, but perhaps more impactfully through the addition of visual similarity-based clustering. With the Sim-to-Sim Mars experiment demonstrating YOCO's ability to detect larger objects such as \textit{sand} and \textit{mountains}, these examples show the direct transfer of this ability from simulation into the real-world through domain adaptation. This provides evidence that the domain adaptive training procedure is successfully forcing the model to extract robust feature representations along with the visual similarity-based clustering procedure having a strong impact towards the recognition success of this type of terrain.

A second observation is that both models seem to favor the \textit{crater} class over the others. This is most certainly due to class imbalance in the training data. From the generated Martian terrain images, craters make up roughly 66\% of the labelled object instances. Even so, YOCO's ability to detect craters outperforms YOLOv3's which is highlighted throughout the qualitative examples. This is clearly demonstrated from the first example in~\autoref{fig:m2020_quals} (top row, left), where we can see an overall improvement in number of craters detected (8 vs 18, despite one misclassification) in a very crater-full image. This pattern continues to be displayed throughout multiple examples, as YOCO detects more craters than YOLOv3, and does not display nearly as many misclassifications as YOLOv3. Further, YOLOv3 demonstrated very few detections of \textit{sand} or \textit{mountains}, where as YOCO is able to detect these classes. YOCO correctly displays cases of sand detection throughout all examples, while YOLOv3 only detects one instance in the third example of~\autoref{fig:m2020_quals2} (bottom row, left). Even with the \textit{mountain} class being visually more challenging to detect, YOCO still makes a good effort to detect this terrain where YOLOv3 does not. The second example in~\autoref{fig:m2020_quals2} (top row, right) shows two overlapping \textit{mountain} class terrain detections in the upper left hand corner of the image frame. Although not entirely accurate the ability to detect such challenging terrain is demonstrated by YOCO, which is a capability translated from the simulation. This could be improved through a better representation of mountains throughout the simulated data, as YOCO performance is tied directly to the quality of the synthetic source domain data.

\subsubsection{\underline{HiRISE Landmarks -}}
\begin{figure*}
\centering

\subfloat{\includegraphics[width=0.24\linewidth]{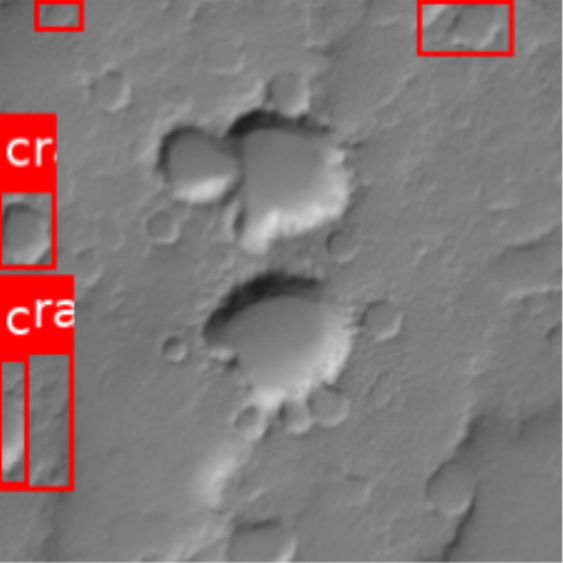}}
\hspace{1mm}%
\subfloat{\includegraphics[width=0.24\linewidth]{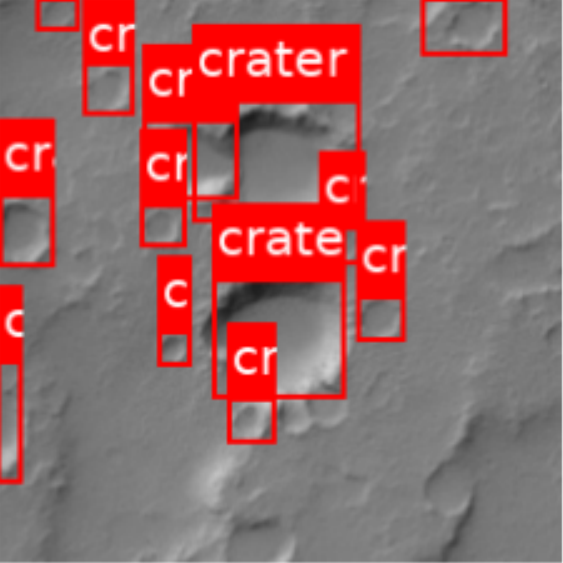}}
\hspace{5pt}%
\subfloat{\includegraphics[width=0.24\linewidth]{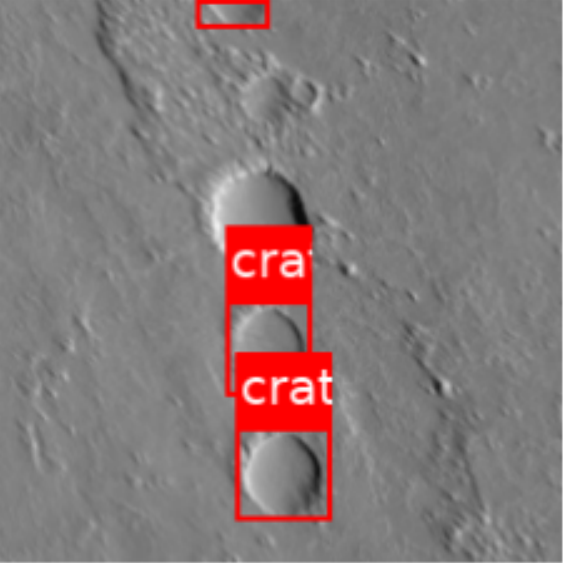}}
\hspace{1mm}%
\subfloat{\includegraphics[width=0.24\linewidth]{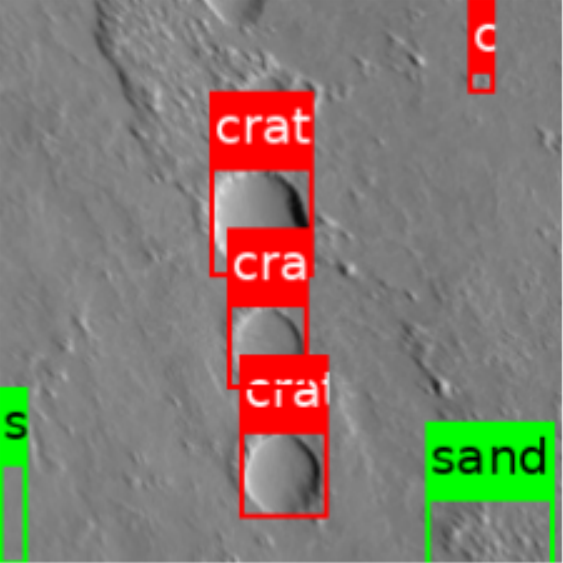}} \\
\vspace{-5pt}

\subfloat{\includegraphics[width=0.24\linewidth]{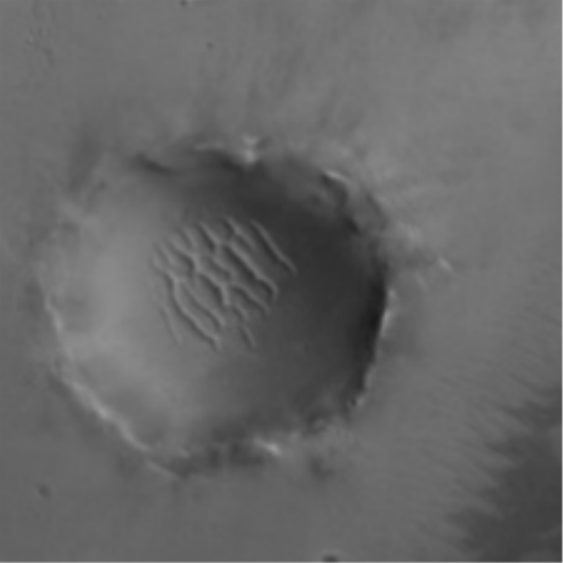}}
\hspace{1mm}%
\subfloat{\includegraphics[width=0.24\linewidth]{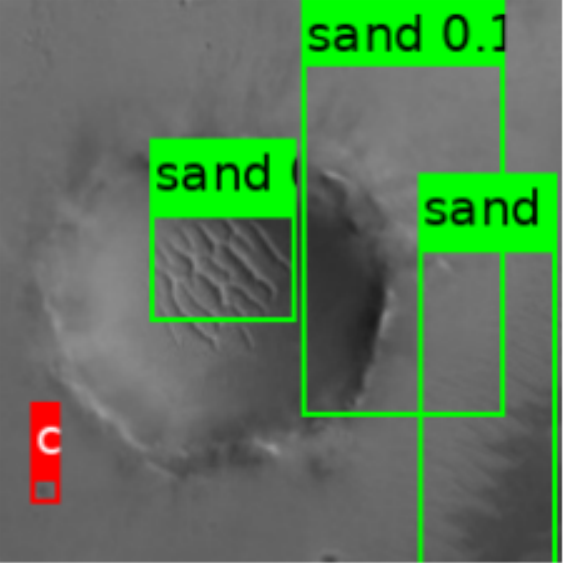}}
\hspace{5pt}%
\subfloat{\includegraphics[width=0.24\linewidth]{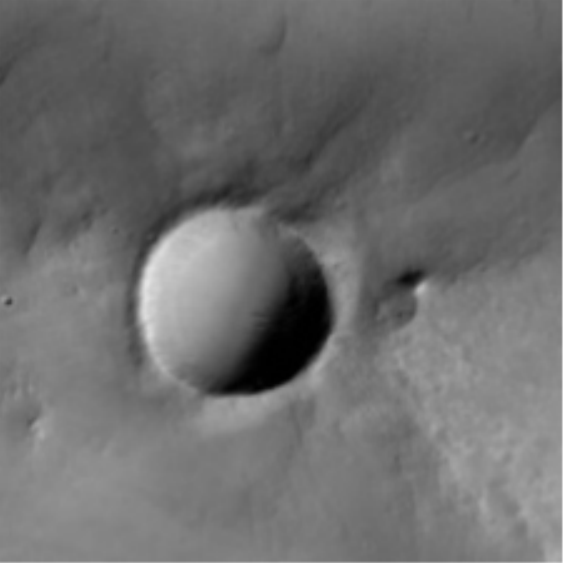}}
\hspace{1mm}%
\subfloat{\includegraphics[width=0.24\linewidth]{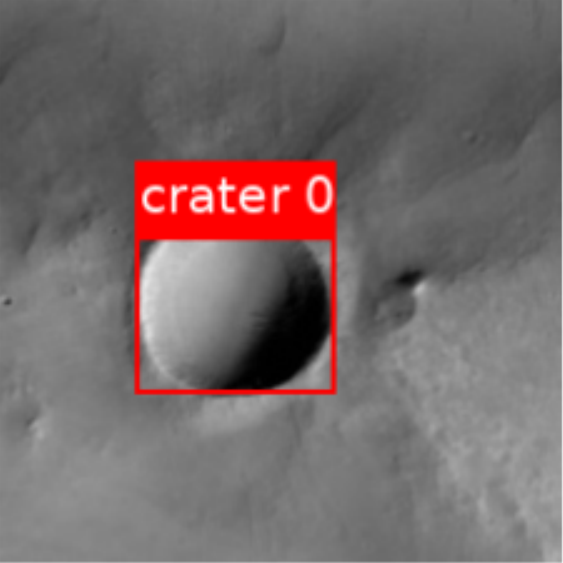}} \\
\vspace{-5pt}

\subfloat{\includegraphics[width=0.24\linewidth]{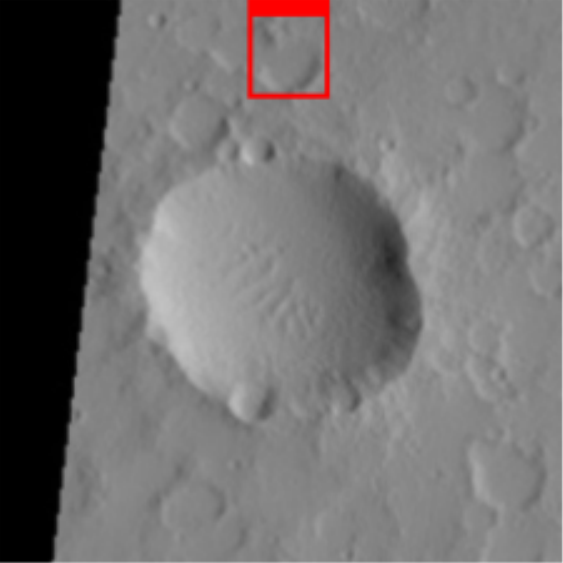}}
\hspace{1mm}%
\subfloat{\includegraphics[width=0.24\linewidth]{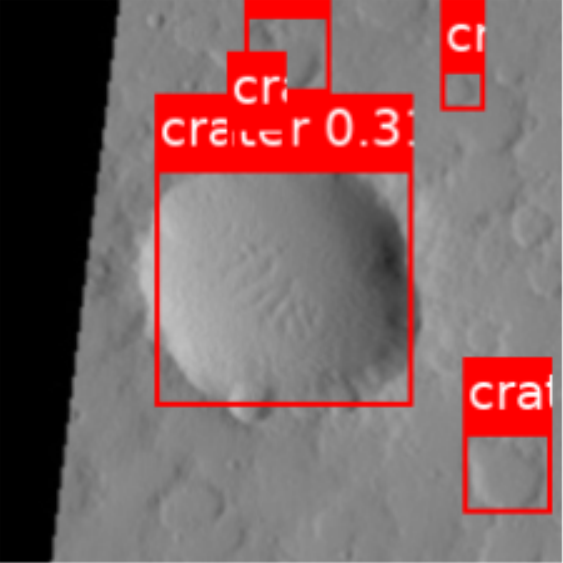}}
\hspace{5pt}%
\subfloat{\includegraphics[width=0.24\linewidth]{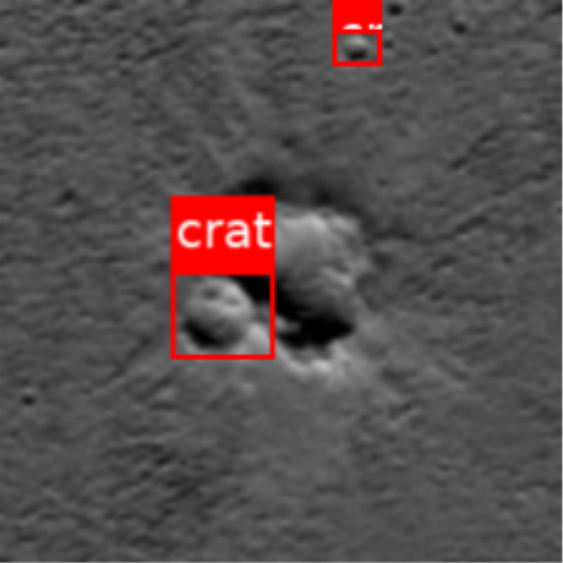}}
\hspace{1mm}%
\subfloat{\includegraphics[width=0.24\linewidth]{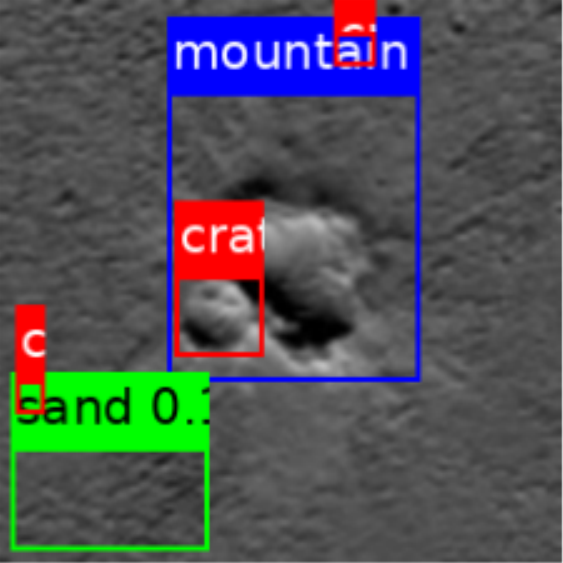}} \\
\vspace{-5pt}

\subfloat{\includegraphics[width=0.24\linewidth]{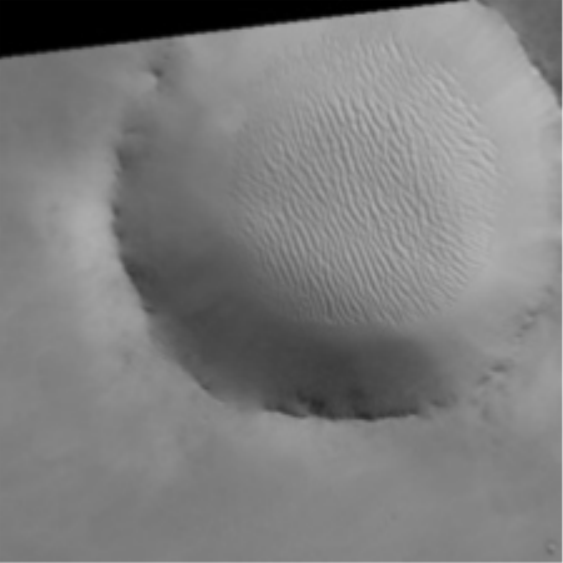}}
\hspace{1mm}%
\subfloat{\includegraphics[width=0.24\linewidth]{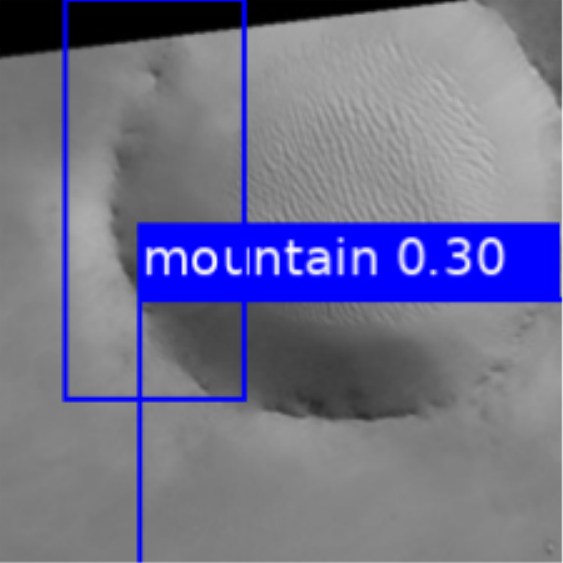}}
\hspace{5pt}%
\subfloat{\includegraphics[width=0.24\linewidth]{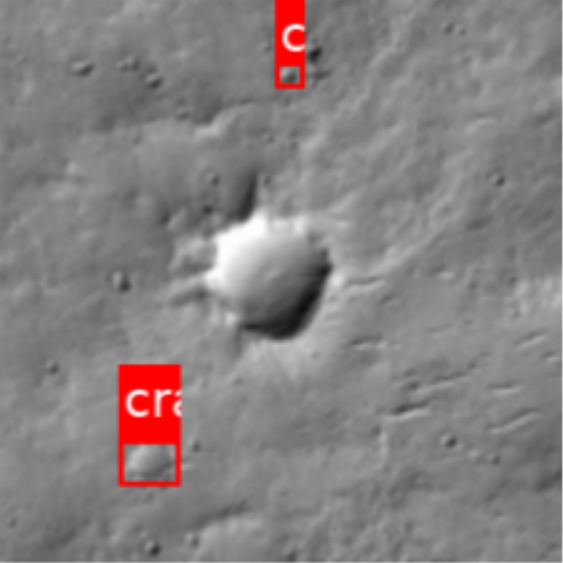}}
\hspace{1mm}%
\subfloat{\includegraphics[width=0.24\linewidth]{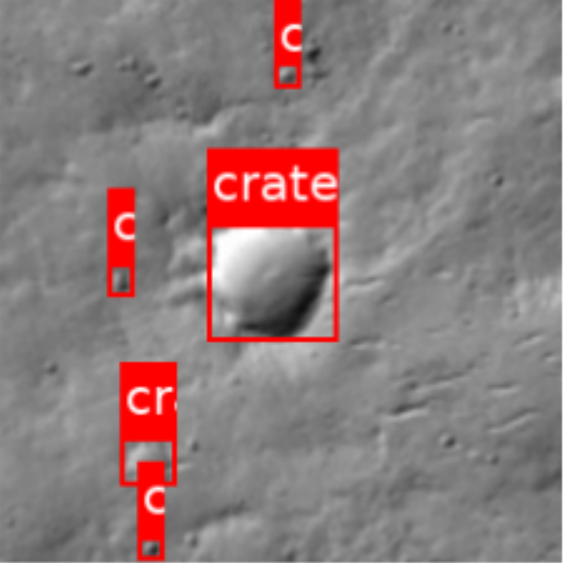}} \\

\caption{Qualitative examples from the Sim-to-Real Mars experiment between YOLOv3 detections (left) and YOCO detections (right) on HiRISE landmark images.}


\label{fig:hirise_quals}
\end{figure*}

This scenario of the experiment aims to qualitate detection performance in an inference domain that's also out of distribution from both source and target data. Various examples of detection performance between YOLOv3 and YOCO on images of cropped Mars landmarks taken from the HiRISE instrument are shown in~\autoref{fig:hirise_quals}. These examples demonstrate the substantially improved performance of YOCO over YOLOv3, even when performing inference on a distribution the model has never seen before. This emulates performance on a more realistic mission concept of operations, as it is more probable to have representative (but not exact) target domain data when training YOCO. We once again see a favor towards crater detection, with a vastly improved number and quality of crater detections coming from YOCO compared to YOLOv3 (clearly demonstrated by the examples in the top row). The more challenging \textit{sand} and \textit{mountain} classes are detected by YOCO in some instances while never being detected by YOLOv3. Although all mountain classifications are formally wrong in each YOCO detection (the actual terrain is crater in each case), the model still detects and localizes what could \textit{plausibly} be a mountain given the scale of terrain feature in the image. This is best highlighted by the example in the bottom left, where the scale of the HiRISE landmark is large enough to cause the ridges of the crater to be detected as mountain. This behavior once again could be fixed through higher fidelity training data in order to teach the model more geometrically correct representations. Nevertheless this example showcases the powerful transfer of representation knowledge between simulated and real-world terrain.

\section{Conclusion}
The ability to consistently and accurately detect instances of hazardous terrain during spacecraft entry, descent, and landing operations is crucial to the assurance of spacecraft safety and mission success. To successfully avoid landing hazards, the spacecraft must be able to find potentially problematic areas of the desired landing site in real-time. With the latest generation of spacecraft compute hardware enabling more advanced algorithms, spacecraft can now leverage terrestrial state-of-the-art systems for the visual detection of landing hazards. To this end we presented \textit{You Only Crash Once} (YOCO), a real-time hazardous terrain detection technique for feature-sparse planetary environments built on the popular YOLOv3 object detection architecture, with simulation-to-real domain adaptation through visual similarity-based clustering that enables training through simulation and removes the need for costly pre-flight reconnaissance mission phases.

Using three sets of experiments we quantitatively and qualitatively demonstrate the effectiveness of our technique. We show improved detection results over YOLOv3 in a standard terrestrial simulation-to-real benchmark, raising both the number and quality of detections. We examine YOCO performance on simulated terrain detections, and use this to apply YOCO on real-world Mars images from two scenarios. Designed for feature-sparse planetary spacecraft landings, we showcase qualitative YOCO detection examples on the landing of the Mars Perseverance Rover as well as orbital landmark images captured by MRO's HiRISE instrument. 




\tiny
\bibliographystyle{AAS_publication}   
\bibliography{references}   
\end{document}